\documentclass[manuscript]{acmart}
\usepackage{cleveref}
\usepackage{dsfont}
\usepackage[T1]{fontenc}
\usepackage{algorithm2e}
\usepackage{algpseudocode}
\usepackage[section]{placeins}
\usepackage{enumitem}
\AtBeginDocument{%
  }

\SetKwComment{Comment}{/* }{ */}

\definecolor{baseline}{RGB}{102,166,30}
\definecolor{benchmark}{RGB}{31,32,65}
\definecolor{baselinehuman}{RGB}{231,41,138}
\definecolor{misreliant}{RGB}{117,112,179}
\definecolor{behavioral}{RGB}{217,95,2}

\newtheorem{definition}{Definition}

\newcommand{\rdecision}{\textsc{R}}

\newcommand{\rprior}{\textsc{R}_{\varnothing}}

\newcommand{\rmis}{\textsc{R}^{\text{m}}}

\newcommand{\baction}{\textsc{B}}

\newcommand{\payoffstate}{\theta}

\newcommand{\binadoptionstate}{\hat{\payoffstate}}
\newcommand{\binadoptionstatespace}{\hat{\Theta}}

\newcommand{\joint}{\pi}

\newcommand{\action}{a}

\newcommand{\binadoptionaction}{\hat{\action}}

\newcommand{\signal}{v}
\newcommand{\signalspace}{V}

\newcommand{\binadoptionsignal}{\hat{\signal}}
\newcommand{\binadoptionsignalspace}{\hat{\signalspace}}

\newcommand{\score}{S}
\newcommand{\proper}{\hat{\score}}

\newcommand{\infoval}{\Delta}

\newcommand{\explanation}{e}

\usepackage{ifthen}

\newcommand{\prob}[2][]{\text{\bf Pr}\ifthenelse{\not\equal{}{#1}}{_{#1}}{}\![{\def\givenn{\middle|}#2}]}
\newcommand{\expect}[2][]{\text{\bf E}\ifthenelse{\not\equal{}{#1}}{_{#1}}{}\![{\def\givenn{\middle|}#2}]}

\newcommand{\featurespace}{X}
\newcommand{\predspace}{Y}

\newcommand{\feature}{x}
\newcommand{\pred}{y}

\newcommand{\hpred}{y^{H}}
\newcommand{\aipred}{y^{AI}}

\newcommand{\actr}{a^{r}}
\newcommand{\actb}{a^{b}}

\DeclareMathOperator*{\argmax}{argmax}

\newcommand{\rl}{\gamma}

\newcommand{\rlr}{\rl^{r}}
\newcommand{\rlb}{\rl^{b}}

\begin{document}

\acmYear{2024}\copyrightyear{2024}
\setcopyright{rightsretained}
\acmConference[ACM FAccT '24]{ACM Conference on Fairness, Accountability, and Transparency}{June 3--6, 2024}{Rio de Janeiro, Brazil}
\acmBooktitle{ACM Conference on Fairness, Accountability, and Transparency (ACM FAccT '24), June 3--6, 2024, Rio de Janeiro, Brazil}
\acmDOI{10.1145/3630106.3658901}
\acmISBN{979-8-4007-0450-5/24/06}


\title{A Decision Theoretic Framework for Measuring AI Reliance}

\author{Ziyang Guo}
\orcid{0009-0004-4200-6774}
\affiliation{%
  \institution{Northwestern Univerisity}
  \streetaddress{633 Clark Street}
  \city{Evanston}
  \state{Illinois}
  \country{USA}
  \postcode{60208}
}
\email{ziyang.guo@northwestern.edu}

\author{Yifan wu}
\affiliation{%
  \institution{Northwestern Univerisity}
  \streetaddress{633 Clark Street}
  \city{Evanston}
  \state{Illinois}
  \country{USA}
  \postcode{60208}
}
\email{yifan.wu@u.northwestern.edu}

\author{Jason Hartline}
\affiliation{%
  \institution{Northwestern Univerisity}
  \streetaddress{633 Clark Street}
  \city{Evanston}
  \state{Illinois}
  \country{USA}
  \postcode{60208}
}
\email{hartline@northwestern.edu}

\author{Jessica Hullman}
\affiliation{%
  \institution{Northwestern Univerisity}
  \streetaddress{633 Clark Street}
  \city{Evanston}
  \state{Illinois}
  \country{USA}
  \postcode{60208}
}
\email{jhullman@northwestern.edu}

\renewcommand{\shortauthors}{Guo et al.}


\begin{abstract}
Humans frequently make decisions with the aid of artificially intelligent (AI) systems. A common pattern is for the AI to recommend an action to the human who retains control over the final decision.
Researchers have identified ensuring that a human has appropriate reliance on an AI as a critical component of achieving complementary performance.
We argue that the current definition of appropriate reliance used in such research lacks formal statistical grounding and can lead to contradictions.
We propose a formal definition of reliance, based on statistical decision theory, which separates the concepts of reliance as the probability the decision-maker follows the AI's recommendation from challenges a human may face in differentiating the signals and forming accurate beliefs about the situation.
Our definition gives rise to a framework that can be used to guide the design and interpretation of studies on human-AI complementarity and reliance.
Using recent AI-advised decision making studies from literature, we demonstrate how our framework can be used to separate the loss due to mis-reliance from the loss due to not accurately differentiating the signals.
We evaluate these losses by comparing to a baseline and a benchmark for complementary performance defined by the expected payoff achieved by a rational decision-maker facing the same decision task as the behavioral decision-makers.
\end{abstract}



\received{20 February 2007}
\received[revised]{12 March 2009}
\received[accepted]{5 June 2009}

\maketitle

\section{Introduction}

AI-advised decision making, in which a human decision-maker has access to the recommendation of an artificial intelligence (AI system) and can choose whether or not to follow it, is often preferred as a means of retaining human control~\cite{bansal2021does} in deploying predictive models.
The motivation behind this approach is \textit{complementary performance}; i.e., the human-AI team can outperform the AI or the human alone.
However, many studies have shown that human-AI teams under-perform the AI alone in tasks~\cite{bansal2021does, buccinca2020proxy, bussone2015role, green2019principles, jacobs2021machine, lai2019human, vaccaro2019effects, kononenko2001machine}. 
One solution to this problem is to identify ways to ensure that the human, as the final decision-maker, has \textit{appropriate reliance} on AI, which is typically defined as submitting the AI recommendation when it is correct and not submitting it when it is not correct.

We argue that this definition of reliance lacks formal statistical grounding, leading to contradictions.
For example, situations in which a human-AI team outperforms the human alone but underperforms the AI alone suggest that the human underrelies on the AI~\cite{bansal2021does}.
However, when researchers apply the above definition of appropriate reliance to their experimental results, they discover that the primary source of performance loss stems from the humans accepting the AI's inaccurate recommendations~\cite{bussone2015role, jacobs2021machine, lai2019human}, considered over-reliance by the conventional definition.

Implicit in discussions of complementarity are assumptions of a human with some internal model of the data-generating process and an AI with its own model. Studying reliance implies that the human consults the AI recommendation, infers the probability that its decision is correct, then decides whether it is worth following its recommendation. 
Problems arise because defining appropriate reliance as submitting the AI's recommendation when it is correct and rejecting it when it is not confounds two challenges a human may face in an AI-advised decision-making: that of forming correct beliefs about the probability that the AI is correct (i.e., prior beliefs), and that of making the optimal decision about whether to follow the AI conditional on the observed information (i.e., posterior beliefs). 
Without a definition that allows separation of different sources of performance loss, the analysis might misinterpret the reasons behind seemingly poor experiment results, leading researchers to prioritize less directly relevant follow-up actions for improving the team.
For example, if the human has inaccurate beliefs about the probability that the AI is correct, this might stem from a lack of information about the prior probability that the AI is correct (potentially addressable by providing the AI's accuracy on held-out data~\cite{yin2019understanding}).
If the human correctly forms the prior belief on AI's accuracy, but fails to correctly choose the AI's recommendation instead of following their own decisions, then this might stem from a lack of motivation to process the information (potentially fixable via cognitive forcing functions~\cite{buccinca2021trust,fogliato2021impact}, better design of incentivations~\cite{li2022optimization}, or better information display such as explantions~\cite{bansal2021does}).

Another issue with the conventional definition of appropriate reliance is that it is a binary measure.
Consequently, researchers cannot distinguish whether the human decision-maker mistakenly used (or did not use) the AI's recommendation in a situation where (A) the probability that relying on their own judgment would have been correct is similar to the probability that the AI was correct versus (B) very different. Intuitively,
over-reliance is a bigger concern in B than in case A. 
We argue that the concept of reliance 
should be characterized within a continuous payoff space to allow for more fine-grained assessment.

We propose a formal definition of AI reliance. 
Following previous work on generating benchmarks for studies of information displays~\cite{wu2023rational}, our approach is grounded in statistical decision theory.
Our definition separates the concepts of a reliance level (the probability that the human decision-maker goes with the AI recommendation) from the belief updating that a rational decision-maker is expected to do upon viewing an instance and associated AI recommendation.
The framework evaluates these two factors of error within the payoff space (i.e.,how much gain can be achieved if the error is corrected) by introducing a benchmark for the best-attainable performance of human-AI collaboration and a baseline for the performance with no collaboration.
We apply the framework to three well-regarded AI-advised decision making experiments from literature~\cite{bansal2021does, lai2019human, fogliato2021impact}.
In all three cases, we show 1) that examining the results against the baseline and benchmark for complementary performance better reveals the limits of human behavioral performance and 2) specific sources of behavioral loss that help explain the experiment results but were not accounted for by the original interpretations of the results.

\section{Formulating Assumptions for Studying Reliance}
In AI-advised decision-making scenarios~\cite{bansal2019updates, wang2019designing}, the human makes a decision about a set of instances with the assistance of an AI recommendation.
In formulating our definition of reliance below, we make several assumptions about this scenario:
\begin{enumerate}[wide, labelindent=0pt,nosep]
    \item The human has an inner model of the data-generating process that enables them to make their own prediction about each instance.
    \item The human consults the AI recommendation prior to making their decision.
\end{enumerate}
There are two benefits to making these assumptions for AI-advised decision-making experiments.
First, the assumptions ensure that participants neither anchor solely on the AI recommendations (completely neglecting to consider their own predictions) nor that they neglect to consult the AI recommendation at all~\cite{buccinca2021trust, fogliato2021impact}. It is difficult to conceive of reliance without such assumptions.
Second, and most importantly, by assuming we have access to the human's own prediction prior to their interaction with the AI recommendation, we can compare the results of experiments we run to a benchmark of complementary performance, which is attained by optimally combining the information contained in the human's predictions with that contained in the AI's recommendations, and a baseline of using either the AI or human only.
We use human recommendation to refer to the human prediction prior to  interaction with the AI recommendation.
\section{Definition of Reliance}

We  define \textit{appropriate reliance}, \textit{over-reliance}, and \textit{under-reliance} on AI recommendations in AI-advised decision making.
Our framework conceives of three roles in the decision problem: a human recommender, an AI recommender, and a decision-maker. 
The two recommenders provide informational input to the decision-maker in the form of recommendations. The decision-maker chooses which recommender to follow on a decision task after seeing the recommendations.

To formalize a decision task requires five key elements (\Cref{tab:notation}): payoff-related states on which the decision is evaluated, a data genertaing model that generates the states and signals that inform about the state, the action, the information (i.e.\ signal) given to the decision-maker,
and a scoring rule  assessing the choice of action under the payoff-related state.


\begin{table*}[htbp]
\small
    \centering
    \begin{tabular}{c|l|l}
    \hline
    & The original decision task &The derived binary-adoption decision task \\
    \hline
     Payoff-related state    &  $\payoffstate = \text{Ground truth } \pred \in \predspace $  & $\binadoptionstate = (\pred, \hpred, \aipred)$ \\
 & & Ground truth $\pred\in \predspace$ \\
 & &Human recommendation $\hpred \in \predspace$ \\
 & &AI recommendation $\aipred \in \predspace$\\
     \hline
     Data generating model    
  & \multicolumn{2}{l}{Feature values $\feature$ from feature space $\featurespace$}\\
 &\multicolumn{2}{l}{$(\feature, \pred) \sim \joint(\featurespace\times\predspace)$}\\
 &\multicolumn{2}{l}{Human recommendation $\hpred$ and AI recommendation $\aipred$:}\\
 &\multicolumn{2}{l}{\quad $(\feature, \hpred) \sim \joint^{H}(\featurespace\times\predspace)$}\\
 &\multicolumn{2}{l}{\quad $(\feature, \aipred) \sim \joint^{AI}(\featurespace\times\predspace)$}\\
         \hline
         Action (choice) & $\action\in \predspace$ &$\binadoptionaction\in \{0 = \text{human}, 1 = \text{AI}\}$\\
     \hline 
     Signal & $\signal = \{\feature, \aipred\}$ & $\binadoptionsignal = \{\feature, \hpred, \aipred\}$\\
     \hline
     Scoring rule (payoff) & $\score(\action, \payoffstate)$
      &$\proper(\binadoptionaction, \binadoptionstate)=\score(\hpred, \pred)$ if $\binadoptionaction=\text{human}$\\
 & &$\proper(\binadoptionaction, \binadoptionstate)=\score(\aipred, \pred)$ if $\binadoptionaction=\text{AI}$\\ \hline
    \end{tabular}
    \caption{Notation for original decision task and derived binary-adoption decision task in our framework.}
    \normalsize
    \label{tab:notation}
\end{table*}

We define the reliance level of an decision-maker on the AI as the overall probability that she chooses the AI recommendation, conditional on the decision maker facing different recommendations from the human and the AI. 
The definition targets a conditional probability, because the reliance level cannot be defined when the human makes the same recommendation as the AI.

\begin{definition}[Reliance]
    The reliance level $\rl$  of any decision-maker on the AI is defined as the conditional probability $\rl=\Pr[\action=\aipred | \aipred\neq \hpred]$ that the decision-maker chooses the AI recommendation,  conditional on the AI recommendation $\aipred$ being different from the human recommendation $\hpred$.
\end{definition}

\subsection{Rational Decision-Maker}
We define the rational decision-maker in a binary-adoption decision task (Table \ref{tab:notation}) derived from the original one.
This derived decision task limits the rational decision-maker to making a final decision by selecting between the human recommendation and the AI recommendation. 
We define the rational benchmark representing the expected performance of a rational Bayesian decision-maker who 
perfectly perceives the provided information and chooses the optimal action under the scoring rule for each decision task \footnote{The expected performance of a rational decision-maker who has access to $\hpred$, $\aipred$, and $\feature$ is an upperbound on the behavioral decision-maker. Though some explanations of $\aipred$ may also be available to behavioral decision-maker, they are a garbling of $\feature$ and $\aipred$ and won't change the rational decision-maker's action.}. 
We use hat symbols to denote the components of the derived decision task, e.g., $\binadoptionstate$ is the payoff-related state in the derived decision task.
The rational benchmark is the maximum payoff that can be expected from a behavioral decision-maker, i.e., the benchmark for complementary performance.
Following the framework proposed by Wu et al. \cite{wu2023rational}, we also define a baseline for expected performance using this rational Bayesian decision-maker.
The rational baseline is the maximum payoff that can be expected from the behavioral decision-maker when they must choose between always going with either the AI or the human recommender, i.e., they do not consult the individual signals in making their decisions.
The rational baseline represents the minimum threshold for achieving complementary performance, 
i.e., the baseline for complementary performance.
Using the rational benchmark and the rational baseline, we define the value of rational complementation, representing the expected improvement in payoff to a rational decision-maker that the joint human+AI setting provides over the better of either the AI or the human alone. 

These three values construct a space of payoffs within which behavioral participants' performance can be quantified and compared.
The rational benchmark also describes the appropriate reliance level, which maximizes the expected payoff.
Throughout the paper, we use superscript $r$ to denote notation for the rational decision-maker. For example, $\actr$ is the action taken by the rational decision maker, and $\rlr$ the rational decision-maker's reliance level. 

\begin{itemize}[wide, nosep]
    \item \textbf{Rational Baseline},
     The rational baseline is the expected performance of the rational decision-maker without access to the signal on a randomly chosen decision task from the experiment.
     Without access to the signal, the rational decision-maker can only make decisions with prior beliefs based on her knowledge of the data-generating model and decision task. 
     This is the better of the two scores achieved by the human alone and the AI alone. 

    \vspace{-4mm}
       \begin{equation*}
    \label{eq:rprior}
    \rprior =\max_{\binadoptionaction} \expect[\joint(\binadoptionstate)]{\proper(\binadoptionaction,\binadoptionstate)} = \max_{\binadoptionaction} \expect[\joint(\binadoptionstate)]{\score(\pred{}  ^{\binadoptionaction},\pred)}.
       \end{equation*}
     
       
    \item \textbf{Rational Benchmark},
    The rational benchmark is the expected performance of the rational decision-maker with the signal on a randomly chosen decision task from the experiment.
    Let $\actr(\binadoptionsignal)$ be the action taken by the rational decision-maker given signal $\binadoptionsignal$. She chooses $\actr$ to maximize her expected utility with  $\joint(\binadoptionstate|\binadoptionsignal)$, the distribution of the payoff-related state conditioned on the signal $\binadoptionsignal$:
           \begin{equation*}
    \label{eq:rposterior}
    \rdecision =\max_{\actr(\cdot)} \expect[\joint(\binadoptionsignal, \binadoptionstate)]{\proper(\actr(\binadoptionsignal), \binadoptionstate)}.
       \end{equation*}
The rational benchmark upperbounds the expected performance of any behavioral decision-maker in the experiment.

\item \textbf{Value of rational complementation},
The value of rational complementation is the increase in payoff over the rational baseline when the rational decision-maker sees the signal.

\begin{equation*}
    \infoval = \rdecision - \rprior.
\end{equation*}

\end{itemize}

The value of rational complementation provides a scale for comparing expected performance in terms of the ``lift'' we see from having access to the information in the signals. 
In the context of AI-advised decision making, it also represents the maximum improvement of performance we can expect from a complementation of the human and the AI conditioned on the information structure of the signals. 
If we treat $\infoval$ as a comparative unit by normalizing all scores within the range where the baseline $\rprior$ is $0$ and the benchmark $\rdecision$ is $1$, we get a sense of the proportion of possible score increase that different settings provide. For example, we could compare expected human performances $\baction_\alpha$ and $\baction_\beta$ under two conditions  $\alpha$ and $\beta$ (e.g., $\alpha$ explanation and $\beta$ explanation) by calculating $(\baction_\alpha - \baction_\beta) / \infoval$.

Given the definitions above, we can define the appropriate reliance level as the reliance level of the rational decision-maker, conditional on the human recommendation being different from the AI recommendation, $\hpred\neq \aipred$. Note that the appropriate reliance level maximizes the expected score of the decision.   

\begin{definition}
    The \textbf{appropriate reliance level} $\rlr$ is the rational decision-maker's reliance level on the AI, $\rlr=\Pr[\actr=1 | \aipred\neq \hpred]$.  
\end{definition}

\subsection{Behavioral Decision-Maker}

    The behavioral decision-maker who completes the decision task takes action $\actb$, and is evaluated by their expected performance on the task. We view the behavioral action as a random variable correlated with the signal, and hence also with the ground truth. Denote the joint distribution as $\joint(\signal, \actb, \payoffstate)$.
    
\begin{itemize}[wide, nosep]
    \item \textbf{Behavioral Performance}
    \begin{equation*}
        \baction = \expect[\joint(\signal, \actb, \payoffstate)]{\score(\actb, \payoffstate)}.
    \end{equation*}
\end{itemize}


We define behavioral \textit{under-reliance} and \textit{over-reliance} by comparing behavioral reliance level $\rlb$ to the appropriate reliance level $\rlr$.

\begin{definition}
    When $\rlb<\rlr$, the behavioral decision-maker \textbf{under-relies} on the AI.
\end{definition}

\begin{definition}
    When $\rlb>\rlr$, the behavioral decision-maker \textbf{over-relies} on the AI. 
\end{definition}


In addition to the reliance level, we analyze the difference between the behavioral decision-maker's expected score and the rational decision-maker's expected score to measure decision quality. To understand why we analyze the difference in score versus in the action space, consider the extreme case where the human recommender and the AI recommender are both uninformative about the ground truth. 
Adopting either the AI recommendation or the human recommendation would achieve an equally bad expected payoff, such that any reliance level between $0\%$ and $100\%$ would perform similarly.
Simply evaluating the reliance level by comparing to the best reliance level ignores the close payoffs achieved by all reliance levels and leads to misleading conclusions.

We separate the behavioral decision-maker's loss in score into two sources: loss from mis-reliance, and what we term discrimination loss, referring to the loss from not accurately distinguishing when the AI recommender has better expected payoff than the human recommender or vice versa.
To separate these sources of loss, we define another benchmark representing the expected score of a rational decision-maker who is constrained to a specific reliance level. 

 \begin{itemize}[wide, nosep]
     \item \textbf{Mis-Reliant Rational benchmark}
     The expected score of a rational decision-maker with reliance level $\rl$:
     \begin{align*}
         \rmis(\rl) = \max_{\actr(\cdot)}\quad&\expect[\joint(\binadoptionsignal, \binadoptionstate)]{\proper(\actr(\binadoptionsignal), \binadoptionstate)}\\
         \text{s.t.}\quad &\Pr[\actr=1 | \aipred\neq \hpred] = \rl \nonumber
     \end{align*}

 \end{itemize}
Hence, the mis-reliant rational benchmark $\rmis$ represents the best score an decision-maker with a given reliance level $\rl$ could attain had they perfectly perceived the probability that the AI is correct relative to the probability that the human is correct on every decision task.
By constraining a rational decision-maker to the same reliance level $\rl$ as each corresponding behavioral decision-maker, we can get a rational decision-maker who simulates the reliance level in the decision rule of the behavioral decision-maker but optimally perceives the signal and arrives at the Bayesian posterior beliefs on each instance. 
By comparing the expected score of these rational decision-makers and behavioral decision-makers,  we can distinguish between the following sources of loss:
\begin{itemize}[wide, nosep]
	\item \textbf{Reliance loss}, the loss from over- or under-relying on the AI, defined as $(\rdecision - \rmis)/\infoval$. We measure reliance loss in payoff space rather than assessing the deviation from the optimal reliance level. The latter treats all errors identically, whereas using payoff space accounts for how big an error is in terms of lost payoff. 

	\item \textbf{Discrimination loss}, the loss from not accurately differentiating the instances where the AI is better than the human from the ones where the human is better than the AI, defined as $(\rmis -\baction)/\infoval$. Since $\rmis$ and $\baction$ have the same reliance level and accept the same percentage of AI recommendations, the difference in the decisions of $\rmis$ and the decisions of $\baction$ lies entirely in accepting the AI recommendations at different instances. $\rmis$ always accepts the top $x\%$ AI recommendations ranked by performance advantage over human recommendations, but $\baction$ may not.
\end{itemize}

\begin{figure}
    \centering
    \includegraphics{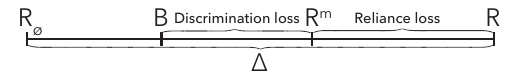}
    \caption{An example of the composition of the quantities defined in our framework. $\rprior$ and $\rdecision$ can be calculated using knowledge of the experiment design, which in our framework includes the human recommendations and the AI recommendations in addition to the components of the decision problem (Table \ref{tab:notation}).
    $\rmis$ and $\baction$ can be calculated given observed data on the human decision-maker's decisions in an AI-assisted scenario.
    }
    \label{fig:losses}
\end{figure}

In other words, we decompose the difference between the best attainable performance in the study ($\rdecision$) and the observed behavior of study participants ($\baction$) into two parts. 
We show an example of the quantities, $\rdecision$, $\rmis$, $\baction$, and $\rprior$, from our framework in Figure \ref{fig:losses}.
Figure \ref{fig:losses} illustrates how the behavioral performance $\baction$ and mis-reliant rational benchmark $\rmis$ are bounded.
$\baction$ must be equal to or lower than the rational benchmark $\rdecision$. 
If $\baction$ is higher than the rational baseline $\rprior$ (i.e., the better performance of either AI recommendations or human recommendations alone), we say $\baction$ fulfills the requirement of complementary performance.
$\rmis$ must fall between $\baction$ and $\rdecision$.


\section{Applying the Framework to AI Reliance Studies}
\label{sec:apply}

We discuss how to apply the framework to AI reliance studies using an example.

\textit{Experiment design and data collection.} 
The first step in applying the framework is to formulate the experiment design as a decision problem by defining the ground truth state, data-generating model, action space, signal, and scoring rule.
Imagine we run an experiment studying AI-advised recidivism decisions with $200$ humans, where each completes 20 trials. In each trial they view a profile of the defendent, and must predict whether the defendent will be re-arrested.
The participants are assisted with an AI model that is deterministic and calibrated on the ground truth. 
We equally divide the $200$ participants into two groups, randomly assigning $100$ to one explanation condition and the other $100$ to a different explanation condition.
All participants first do the $20$ instances by themselves before they see any AI recommendations, then make final decisions on the same $20$ instances with the AI assistance.
For every correct decision on the second batch of trials, the participant receives $\$0.5$ as incentivization.
The decision tasks are formalized in Table \ref{tab:applying_example} in Appendix~\ref{app:decision_task}.
When the experiment is complete, we have collected $4000$ decision observations in total.
Each observation includes information about the profile of the defendent, the outcome of whether the defendent is re-arrested, the human recommendation on the first batch of trials, the AI recommendation, the explanation of the AI recommendation, and the final decision on the second batch of trials.

\textit{Rational baseline $\rprior$.} 
Recall that the rational baseline represents the expected performance of the rational decision-maker without access to the signal on the derived binary-adoption decision task from the experiment. Hence, the best action is the better of always following the AI and always following the human recommendation. We estimate the rational baseline by identifying the best-response to the empirical distribution of states in the $4000$ observations experiment.
This calculation is illustrated in Algorithm~\ref{alg:baseline} in Appendix~\ref{app:algorithms}.


\textit{(Approximating) Rational benchmark $\rdecision$.}
To calculate the rational benchmark we identify the best response to each signal. When the signal space has finite size, we can calculate the rational benchmark by simulating the best response to each signal on the empirical distribution of the experiment observations. However, for a large number of decision tasks in the literature (including, e.g., the demonstrations in \Cref{sec:demonstration}), the signal space has near infinite size (e.g., it involves text documents) such that each experimental observation might involve a different unique signal. Thus, the identified best response action may overfit to the data relative to the true expected score of the rational decision-maker on a randomly chosen decision task from the experiment. 
We approximate the rational benchmark by designing an upperbound and a lowerbound. 
\begin{itemize}[wide, nosep]
    \item Upperbound: Overfitting to the empirical distribution. We calculate the rational benchmark on the empirical joint distribution $\Tilde{\joint}(\binadoptionstate, \binadoptionsignal)$ over the payoff-relevant state $\binadoptionstate$ and the signal $\binadoptionsignal$, treating the empirical distribution as the true data generating model. Algorithm~\ref{alg:overfit} in Appendix~\ref{app:algorithms} calculates this empirical distribution. 
\end{itemize}

To see why this is an upperbound and why we call it overfitting, consider the case where the signal space is continuous. Each entry in the experiment data has a distinct signal. 
Without repetition, it is impossible to approximate the true distribution of the payoff-relevant state $\binadoptionstate$ conditioning on each signal $\binadoptionsignal$. Treating the empirical distribution as the true data generating model, there is no randomness in the payoff-relevant state given the rational decision-maker's knowledge. 
\begin{itemize}[wide, nosep]
    \item Lowerbound: Learning the best response on the optimally discretized empirical distribution to avoid overfitting. Assuming continuity on the joint distribution $\Tilde{\joint}(\binadoptionstate, \binadoptionsignal)$ over the payoff-relevant state $\binadoptionstate$ and the signal $\binadoptionsignal$, we approximate the rational benchmark by coarsening the signal space into finite discrete signals $\Tilde{\binadoptionsignal}_1, \Tilde{\binadoptionsignal}_2, \dots, \Tilde{\binadoptionsignal}_k$,  and calculating the best response on the empirical distribution over the discretized space $\{\Tilde{\binadoptionsignal}_i\}_i$. An example using the $k$-means algorithm to discretize the signals is shown in Algorithm~\ref{alg:kmeans} in Appendix~\ref{app:algorithms}. 
\end{itemize}

To see why this is an lowerbound on the rational benchmark, first note that the rational decision-maker with the true data generating model can always perform the same discretization as the algorithm on the signal space, and such discretization to the signal can only decrease the expected performance.
It remains to make sure the discretization is not too fine, such that the estimate on the empirical distribution is close to the rational decision-maker's expected payoff on the discretized signal (i.e.\ the estimate does not overfit to the data points from the experiment).  We ensure this by performing cross-validation on the estimated average payoff. We randomly split the experiment data into a training set and a test set. Intuitively, increasing the number of clusters $k$ leads to an expected payoff closer to the rational benchmark, but a higher gap between the estimated payoff on the clustering set and the test set (a.k.a.\ the generalization error). We select $k$ to balance the increase in expected payoff and the generalization error.  

The calculation of the rational benchmark hence takes an empirical distribution as input. For a finite signal space, the rational benchmark is calculated on the empirical distribution. For an infinite signal space, the upperbound is calculated on the empirical distribution, while the lowerbound is calculated on the discretized empirical distribution. 
Regardless of which bound we are calculating, given an empirical distribution (e.g, the $4000$ observations), we simulate the rational decision-maker's decision. 
For each observation, the rational decision-maker receives a signal (raw signal or discretized signal) 
and  calculates the posterior distribution of states given the signal by Bayes rule, denoted as $\joint(\binadoptionstate | \binadoptionsignal) = \frac{\joint(\binadoptionstate, \binadoptionsignal)}{\joint(\binadoptionsignal)}$.
We pick the action with higher expected payoff under the posterior distribution on the current observation.
We repeat this process for all observations and then take the expectation on all the rational benchmarks we get.
We can take the conditional expectation across different conditions, e.g., different explanations.
This calculation is illustrated in Algorithm~\ref{alg:benchmark} in Appendix~\ref{app:algorithms}.

\textit{Behavioral performance $\baction$.}
The expected performance of a behavioral decision-maker's final decision
is estimated on the joint behavior of the behavioral decision-makers in the experiment, denoted as $\joint(\signal, \payoffstate, \actb)$.
We can use the observations to directly represent the joint behavior of the behavioral decision-makers or estimate using a model trained on the observations to predict the behavioral decisions\footnote{When we estimate the joint behavior by a model, how good the estimates of behavioral performance are will depend on how well the model predicts the behavioral data.}.
This calculation is illustrated in Algorithm~\ref{alg:behavioral} in Appendix~\ref{app:algorithms}.

\textit{(Approximating) Mis-reliant rational benchmark $\rmis$.}
The mis-reliant rational benchmark is the expected score of a rational decision-maker with the same behavioral reliance level as the human participant. 
To calculate this, we simulate the rational decision-maker completing the same set of trials as the behavioral decision-makers do but additionally constrain the reliance level to be the same as the reliance level produced by the behavioral decision-makers.
In our example experiment, each behavioral decision-maker completes $20$ trials with reliance levels, $\rl^b = \Pr[\actb = \aipred|\aipred \neq \hpred]$.
As the rational decision-maker traverses the $4000$ observations, like behavioral participants she should engage in $20$ consecutive trials for each set. 
Suppose that the signals that the rational decision-maker receives in the $20$ consecutive trials are $\binadoptionsignal_1, \ldots, \binadoptionsignal_{20}$.
For each signal $\binadoptionsignal_i$, the rational decision-maker knows the posterior payoffs, i.e., $\expect[\joint(\binadoptionstate | \binadoptionsignal_i)]{\score(\aipred, \pred)}$ and $\expect[\joint(\binadoptionstate | \binadoptionsignal_i)]{\score(\hpred, \pred)}$.
Then, the rational decision-maker ranks the signals in decreasing order of $\expect[\joint(\binadoptionstate | \binadoptionsignal_i)]{\score(\aipred, \pred)} - \expect[\joint(\binadoptionstate | \binadoptionsignal_i)]{\score(\hpred, \pred)}$ and accepts the AI recommendation from the first signal in the sorted list, up to a $\rl^b$ fraction of $20$ signals. 
We take the expectation over all observations (or conditionally on the manipulated variable of interest depending on the study design). This calculation is illustrated in Algorithm~\ref{alg:misreliant} in Appendix~\ref{app:algorithms}.
Note that estimation of the mis-reliant rational benchmark faces the same risk of overfitting as the rational benchmark. When the signal space is infinite, we approximate the mis-reliant rational benchmark the same way that we do the rational benchmark by calculating the upper- and lower-bound.

\textit{Quantifying uncertainty.}
All the quantities calculated by the above algorithms are point estimates of the expectations.
To get a robust estimate, we bootstrap to compute the expectation.
For each iteration in bootstrapping, we sample 
from the $4000$ observations, and run the four algorithms on the ratio of the sample. 
The estimations of the expected payoff generated through iterations quantify the uncertainty.
This calculation is illustrated in Algorithm~\ref{alg:uncertainty} in Appendix~\ref{app:algorithms}.
\section{Demonstration}

\label{sec:demonstration}

We apply our framework to three AI-advised decision making experiments~\cite{bansal2021does, lai2019human, fogliato2021impact}. \footnote{We use the upper bound (overfit) method to approximate the rational benchmarks and the mis-reliant rational benchmark, i.e., estimating the empirical distribution using the observations of signals and payoff-relevant state and treating the empirical distribution as the true data generating model. We confirmed our conclusions from this approach using the approximation of the rational benchmark with discretized signals in Appendix~\ref{app:demo_test_performance}.}
We reanalyze the reliance levels of behavioral decision-makers within the payoff space by comparing to the rational baseline and the rational benchmark. We also identify the discrimination loss.\footnote{See our supplementary materials for complete analysis with full code and original data: \url{https://osf.io/2cbxf/?view_only=fd9c2e8e1dd24aa787af05dadafe4bcf}}

\subsection{On Human Prediction with Explanations and Predictions of Machine Learning Models~\cite{lai2019human}}

Lai and Tan \cite{lai2019human} compare different approaches to integrate an AI in the task of detecting deception in hotel reviews.
\subsubsection{Experiment design.}
Following \cite{ott2011finding}, participants are asked to look up a hotel review and then make a decision on whether the review is genuine or deceptive.
Lai and Tan \cite{lai2019human} proposed seven conditions with different levels of AI assistance along a hypothesized spectrum from full human agency to full automation: no information from the AI, only example-based explanation, only highlight-feature explanation, only heatmap explanation, only predicted label, predicted label with random heatmap explanation, predicted label with example-based explanation, predicted label with heatmap explanation, and predicted label with accuracy.
Since the reliance problem we study is defined only for the scenario where the AI recommendation is provided to the human decision maker, we analyze only the five conditions including AI information.
The decision task is summarised in Table~\ref{tab:lai_decision} in Appendix~\ref{app:decision_task}.

\subsubsection{Analysis}

\begin{figure*}
    \centering
    \includegraphics[width=\textwidth]{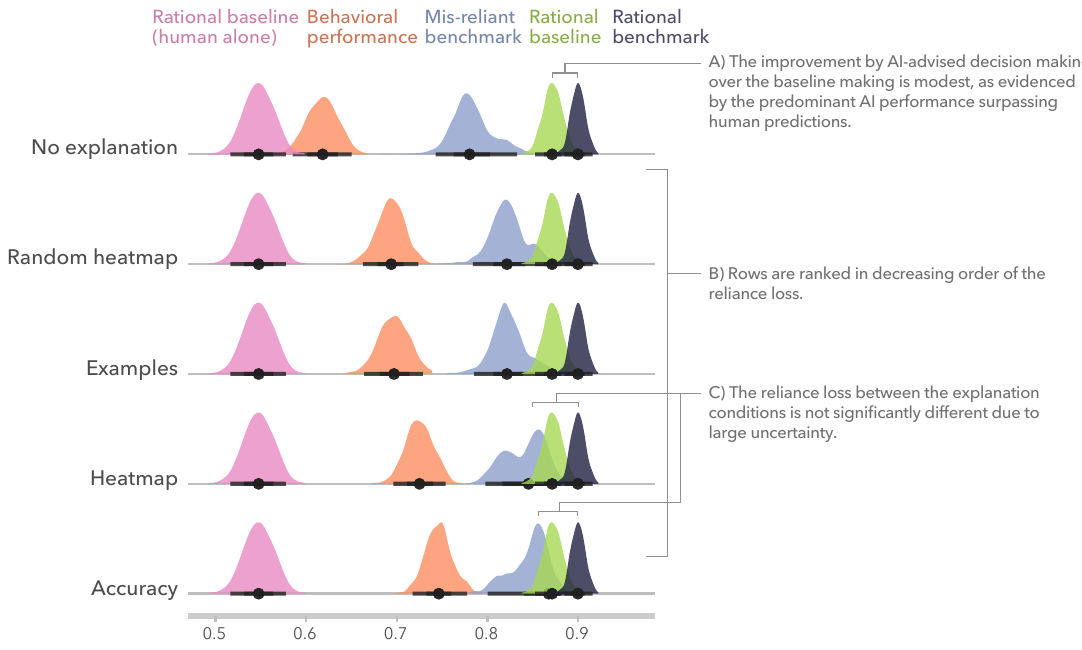}
    \caption{Expected payoffs of benchmarks, baselines, and observed performance in Lai and Tan \cite{lai2019human}.}
    \label{fig:lai_results}
\end{figure*}

\begin{figure}
    \centering
    \includegraphics[width=0.8\linewidth]{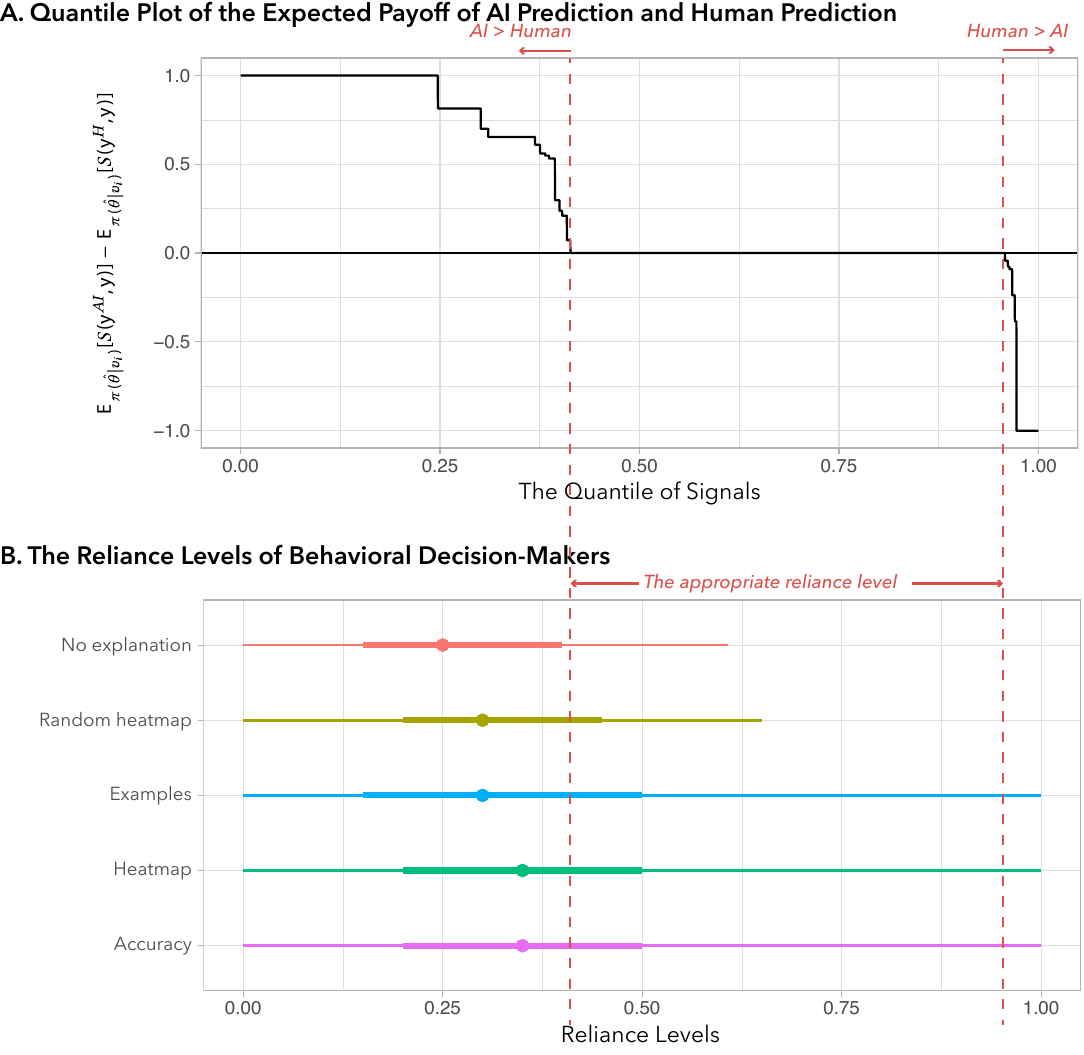}
    \caption{Plots demonstrating how the rational agent arrives at the appropriate reliance level by maximizing her payoff in the decision-making problem defined by Lai and Tan~\cite{lai2019human}, including A) quantile plot (\textit{y-axis}: $\expect[\joint(\binadoptionstate | \signal_i)]{\score(\aipred, \pred)} - \expect[\joint(\binadoptionstate | \signal_i)]{\score(\hpred, \pred)}$ ranked in descending order; \textit{x-axis}: the cummulative probability (quantile) of signal $\signal_i$) and B) 50\% and 95\% intervals on behavioral decision-makers' reliance levels.}
    \label{fig:lai_reliance}
\end{figure}

The conclusions drawn by Lai and Tan \cite{lai2019human} include: AI-advised decisions were better when the AI system interfered more with the human decision-maker's process, and trust in the AI recommendation increased with more AI-based information. Trust was evaluated by the rate at which the AI recommendations were accepted.
Their results ranking the AI-based conditions by both performance and trust is (from worst to best) were: no predicted label < only predicted label < predicted label with random heatmap explanation < predicted label with example-based explanation < predicted label with heatmap explanation < predicted label with accuracy.
Using our approach, we examine the ranking of behavioral performance within the scale created by the rational baseline and rational benchmark. Instead of evaluating reliance as rate of acceptance of AI recommendations, we evaluate the reliance level of the behavioral decision-makers in payoff space.

Extending the author's original conclusions, we find that \textbf{the rational baseline dominates almost all other quantities in our framework except the rational benchmark}, including the behavioral performance and the mis-reliant rational benchmark across all explanation conditions, as shown in Figure~\ref{fig:lai_results} (\textbf{\textcolor{baseline}{the rational baseline}} and \textbf{\textcolor{benchmark}{the rational benchmark}}).
Additionally, \textbf{the rational benchmark only improves marginally over the rational baseline}, i.e., the rational decision-maker does not gain much from access to human recommendations, as shown in Figure~\ref{fig:lai_results}A (\textbf{\textcolor{benchmark}{the rational benchmark}} and \textbf{\textcolor{baseline}{the rational baseline}}).
Consequently, it is hard to expect behavioral decision-makers to achieve complementary performance.
These findings suggest that the experimental design was poorly suited for studying complementary performance, because the AI consistently outperforms the human.

Using our approach, we extend the authors' results by observing that \textbf{different explanation conditions result in different levels of discrimination loss and reliance loss.}
For example, the condition with heatmap explanations and the condition directly providing model accuracy show similar reliance loss (Figure~\ref{fig:lai_results}C) but the discrimination loss in the latter is smaller than the former.
This suggests why showing accuracy can help the behavioral decision-makers achieve higher performance than heatmap explanations: the accuracy information helps the behavioral decision-makers better differentiate instances where the AI predictor outperforms the human predictor from those where the human predictor outperforms the AI predictor, presumably because it provides information on the joint distribution of the AI recommendation and the ground truth that is absent from the heatmap explanations.


\subsection{Does the Whole Exceed its Parts?~\cite{bansal2021does}}

Bansal et al. \cite{bansal2021does} use an online crowdsourced experiment to investigate the effects of explanations on the degree of complementary performance achieved by AI-advised humans.
In contrast to prior studies like \cite{lai2019human}, Bansal et al. \cite{bansal2021does} controlled the AI's accuracy to be comparable to the humans', to avoid the AI being obviously better than human performance on the task. 
\subsubsection{Experiment design}

The experiment compares human-AI team decisions across four approaches to explaining AI recommendations: no explanation, explanation for the most confident AI recommendation, explanations for the top-2 most confident AI recommendations, and adaptively showing explanations for the top-1 or top-2 most confident AI recommendations, randomly assigned between subjects.
The participants are tasked with using the AI recommendation and its explanation for two tasks: sentiment classification and LSAT (multiple-choice questions where one of four choices is the correct answer).
Because the manipulation of interest (explanation types) and conclusions drawn about the complementary performance of the human-AI teams across different explanation types are the same between the two tasks, we analyze only the results of the LSAT task. 
The decision task is summarised in Table~\ref{tab:notation_bansal} in Appendix~\ref{app:decision_task}.
\subsubsection{Analysis}

\begin{figure*}
    \centering
    \includegraphics[width=0.8\textwidth]{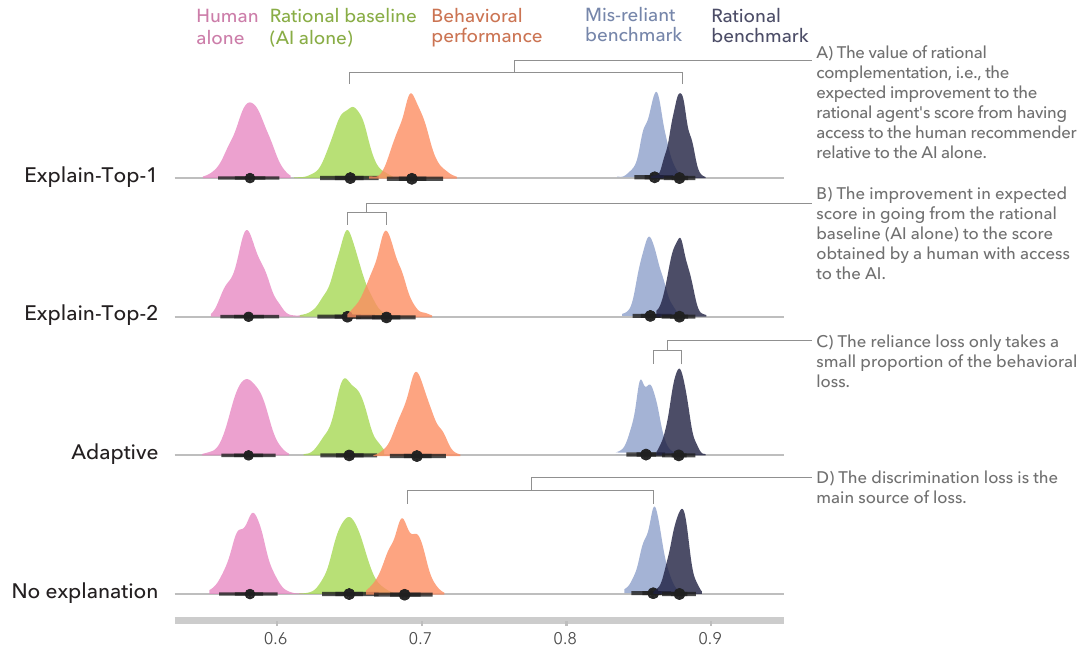}
    \caption{Expected payoffs of benchmarks, baselines, and observed performance in Bansal et al. \cite{bansal2021does}.
    }
    \label{fig:bansal_results}
\end{figure*}

\begin{figure}
    \centering
    \includegraphics[width=0.8\linewidth]{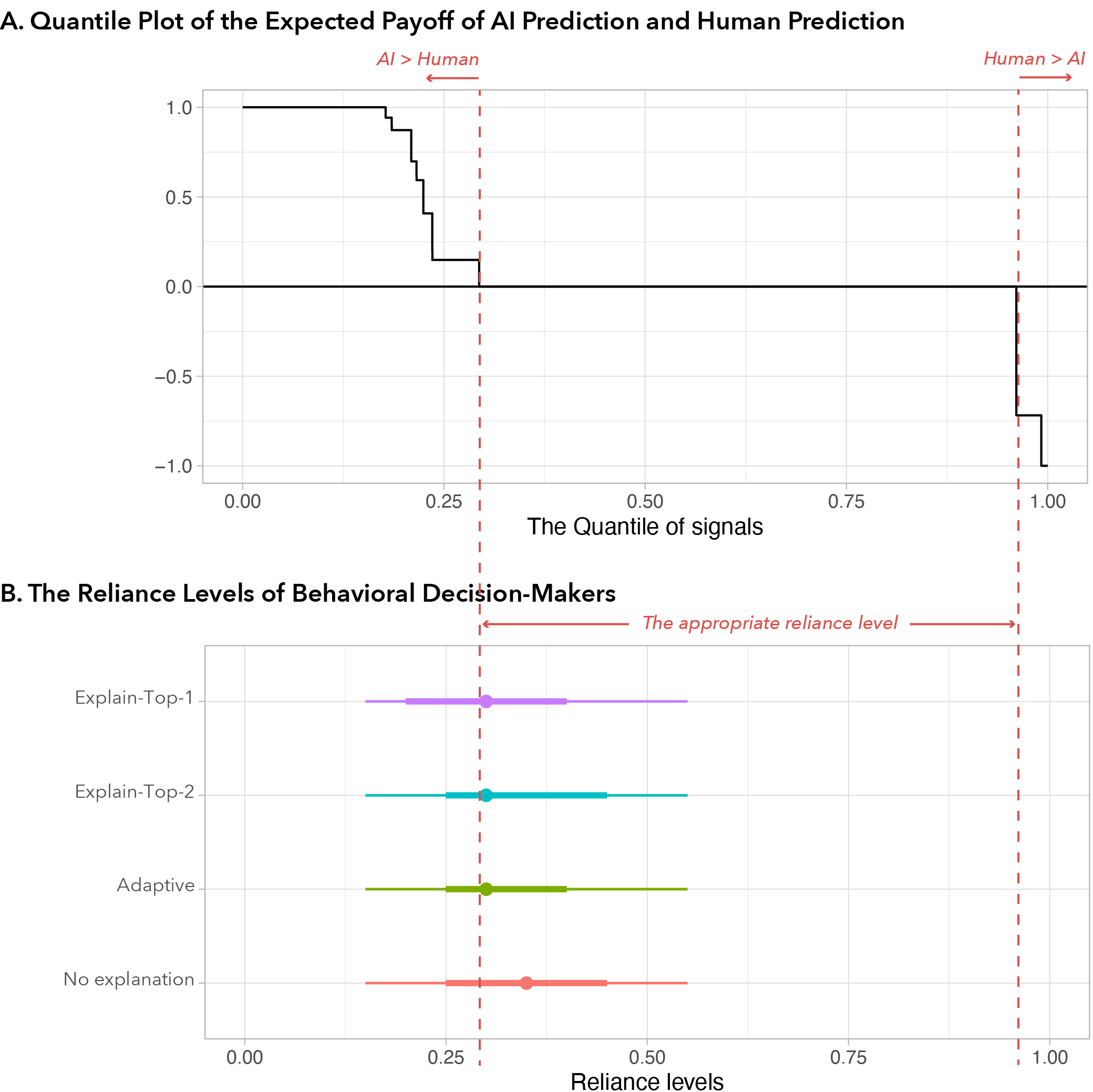}
    \caption{Plots demonstrating how the rational agent arrives at the appropriate reliance level by maximizing her payoff in the decision-making problem defined by Bansal et al.~\cite{bansal2021does}, including A) quantile plot (\textit{y-axis}: $\expect[\joint(\binadoptionstate | \signal_i)]{\score(\aipred, \pred)} - \expect[\joint(\binadoptionstate | \signal_i)]{\score(\hpred, \pred)}$ ranked in descending order; \textit{x-axis}: the cummulative probability (quantile) of signal $\signal_i$) and B) 50\% and 95\% intervals on behavioral decision-makers' reliance levels.}
    \label{fig:bansal_reliance}
\end{figure}

Bansal et al. \cite{bansal2021does} drew several conclusions from their results: AI-advised decision making achieved complementary performance (i.e., a higher payoff than expected of the human or AI alone), 
and presenting explanations to the human-AI team led to no observable performance improvements using null hypothesis significance testing (NHST) with $\alpha=0.05$.
The authors speculated that the reason they did not observe improvement from explanations is because people over-relied on the AI when explanations are provided.
This is supported by evidence that providing explanations increased decision performance when the AI was correct and decrease it when the AI was incorrect.
We use our framework to evaluate this conclusion.
Specifically, we compare the observed behavioral payoffs to the rational baseline and rational benchmark, and evaluate the reliance level of participants in payoff space by comparing the behavioral payoffs to the mis-reliant rational benchmark.
Our results are shown in Figure~\ref{fig:bansal_results}.

Extending the authors' original conclusions, we find that \textbf{despite the behavioral decision-makers achieving complementary performance, there is still considerable room for improvement}, shown as the distance between \textbf{\textcolor{behavioral}{the behavioral performance}} and \textbf{\textcolor{benchmark}{the rational benchmark}} (Figure~\ref{fig:bansal_results}A and B).
The \textbf{\textcolor{behavioral}{behavioral payoff}} surpasses the \textbf{\textcolor{baseline}{rational baseline}}, as shown in all rows representing different explanation conditions in Figure~\ref{fig:bansal_results}. 
This comparison leads to the authors' conclusion that complementary performance is observed in every condition.
However, comparing to the \textbf{\textcolor{benchmark}{rational benchmark}}, the \textbf{\textcolor{behavioral}{behavioral decision-makers}} only improve a small proportion over the \textbf{\textcolor{baseline}{rational baseline}} (Figure~\ref{fig:bansal_results}). 
Our analysis more clearly demonstrates the remaining need to identify ways to bridge the remaining substantial gap.

Applying NHST as in the original study, we corroborate the authors' conclusion that there are \textbf{no significant improvements for explanation conditions over the no explanation condition}. 
Using our approach we confirm there are not significant reductions in either discrimination loss or reliance loss.
For example, in Figure~\ref{fig:bansal_results} (\textbf{\textcolor{behavioral}{behavioral performance}} and \textbf{\textcolor{misreliant}{mis-reliant rational benchmark}}), the behavioral decision-makers in the no explanation and the adaptive explanation condition achieve similar performance; the same is true of the Explain-Top-1 and Explain-Top-2 conditions.

Further extending the original conclusions, 
we find that \textbf{despite the over-reliance shown by the original paper, poor reliance itself is not the main source of loss.}
While the behavioral decision-makers' reliance levels across all conditions \textit{are} higher than the optimal reliance level in expectation represented by the rational benchmark, our analysis suggests that miscalibrated reliance of the behavioral decision-makers does not lead to substantial loss in payoff.
As shown in Figure~\ref{fig:bansal_results}C, \textbf{\textcolor{misreliant}{the mis-reliant rational benchmarks}} across all conditions are very close to \textbf{\textcolor{benchmark}{the rational benchmark}}, such that reliance loss is very minor compared to the total behavioral losses.


Instead our approach shows that \textbf{the behavioral decision-makers have substantially lower performance compared to the rational benchmark due to large discrimination loss} (i.e., accepting the AI recommendations for the wrong instances), as shown in Figure~\ref{fig:bansal_results}D.
Combined with the evidence that the behavioral decision-makers have low reliance loss, this could suggest that the explanations be designed specifically to help users distinguish the intance where the AI is expected to succeed from those where the AI is expected to fail, instead of aiming to calibrate the human's overall trust in the AI's accuracy or adjusting the human's decision rule.
For example, explanations could give information on the joint distribution of AI recommendation and the ground truth, i.e., $\joint(\aipred, \pred)$ rather than focusing on describing only the decision rule of AI, e.g., as in LIME~\cite{ribeiro2016should} or SHAP~\cite{lundberg2017unified}.
\subsection{The Impact of Algorithmic Risk Assessments on Human Predictions and its Analysis via Crowdsourcing Studies~\cite{fogliato2021impact}}

Fogliato et al. \cite{fogliato2021impact} conduct an online crowdsourcing experiment where participants face the task of assessing a defendant's risk of re-arrest after viewing the defendant's profile. 
The experiment investigates the research questions of whether anchoring effects impact participants' recommendations and whether the evaluation of participants' decisions depends on the types of recommendations (probablity or binary decision), both of which can be modeled as decision tasks in our framework.

\subsubsection{Experiment Design}

The experiment compares AI-assisted human recommendations under two different conditions: anchoring and non-anchoring. 
Participants assigned to the anchoring condition see the question presented together with the AI's recommendation, while under the non-anchoring condition, participants are asked to predict the risk before seeing AI recommendation and then to revise their assessment after having AI recommendation. 
In each question, participants are shown the profile of a defendant, including demographics, current charge, and criminal history.  
Participants are asked to report: 1) the probability of the defendant being re-arrested from $[0, 100\%]$, and 2) a binary choice of whether the defendant will be re-arrested within a given duration or not. 
The decision tasks for probabilty and binary decision are summarised in Table \ref{tab:notation_fogliato} in Appendix~\ref{app:decision_task}.

\subsubsection{Analysis}

\begin{figure*}[t]
    \centering
    \includegraphics[width=0.8\textwidth]{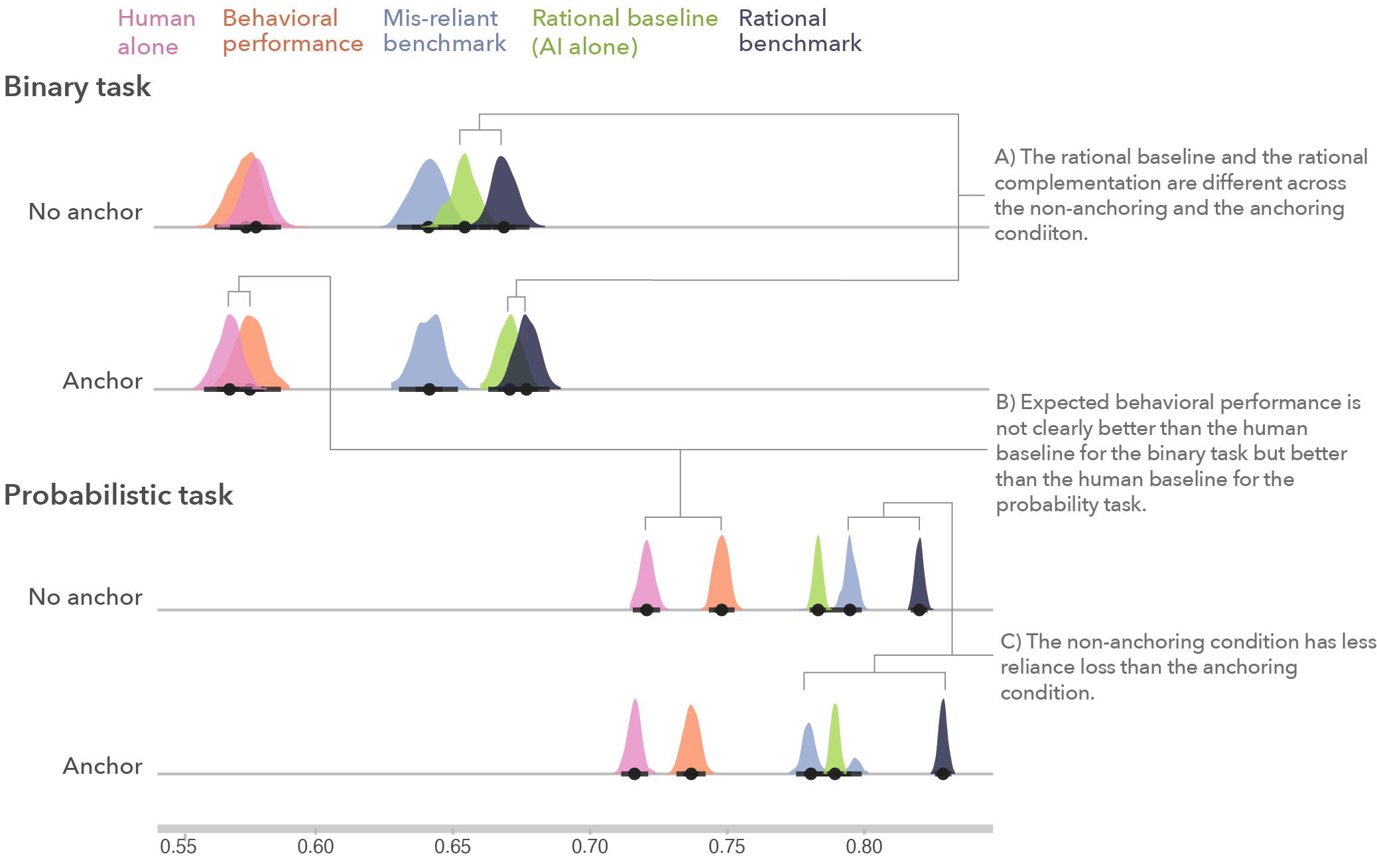}
    \caption{Expected payoffs of benchmarks, baselines, and observed performance in Fogliato et al. \cite{fogliato2021impact}}
    \label{fig:fogliato_results}
\end{figure*}

\begin{figure}
    \centering
    \includegraphics[width=0.8\linewidth]{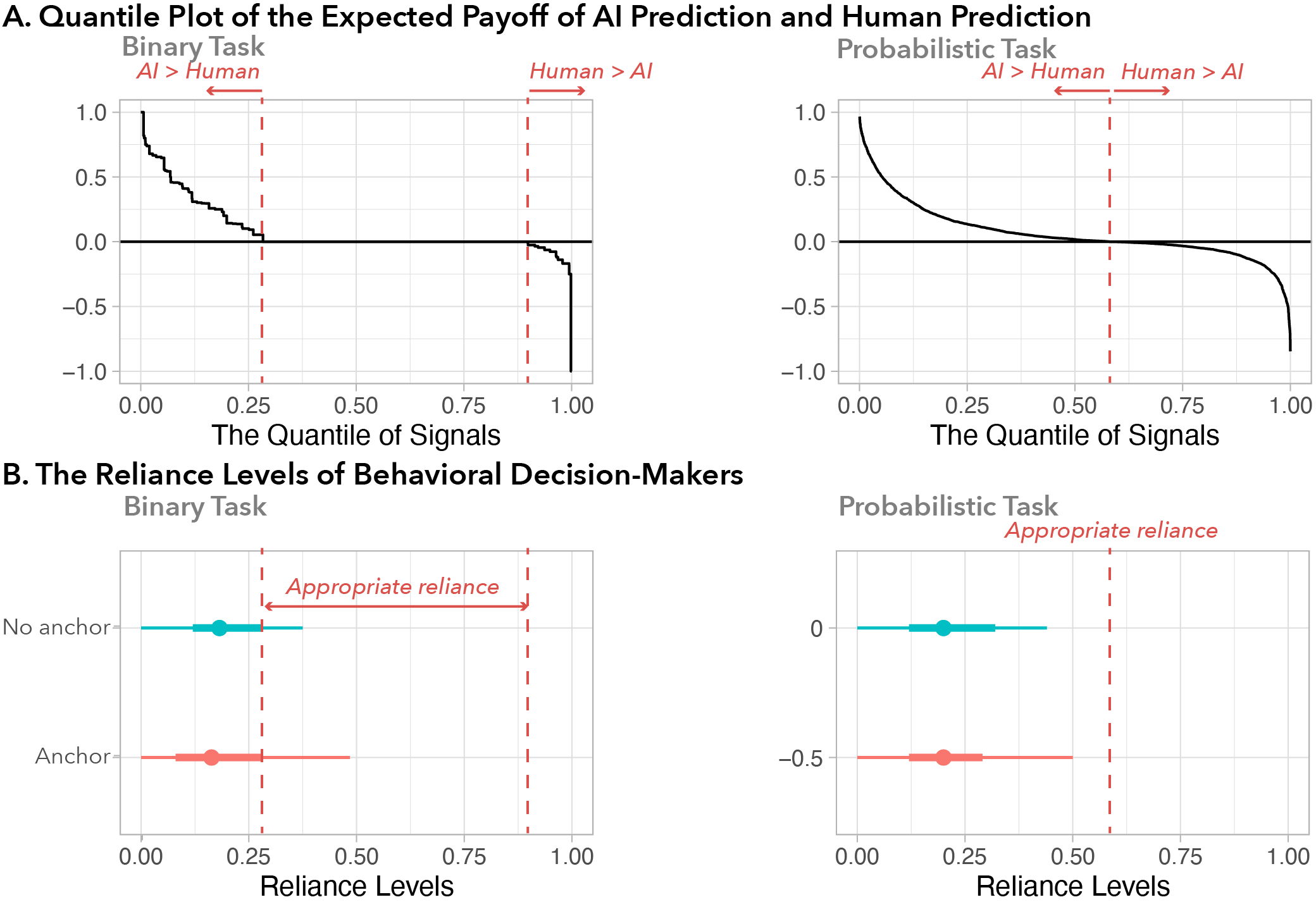}
    \caption{Plots demonstrating how the rational agent arrives at the appropriate reliance level by maximizing her payoff in the decision-making problem defined by Fogliato et al.~\cite{fogliato2021impact}, including A) quantile plot (\textit{y-axis}: $\expect[\joint(\binadoptionstate | \signal_i)]{\score(\aipred, \pred)} - \expect[\joint(\binadoptionstate | \signal_i)]{\score(\hpred, \pred)}$ ranked in descending order; \textit{x-axis}: the cummulative probability (quantile) of signal $\signal_i$) and B) 50\% and 95\% intervals on behavioral decision-makers' reliance levels.}
    \label{fig:fogliato_reliance}
\end{figure}


Fogliato et al. \cite{fogliato2021impact} report that 1) the probability of re-arrest reported by the participants did not uniformly map to their binary decision, such that behavioral predictive performance and reliance level must be considered separately, 
and 2) no clear differences between
participants' accuracy, false positive rate, false negative rate, positive predicted values, or AUC were found between the anchoring and no anchoring condition.
Our analysis of their results is shown in Figure~\ref{fig:fogliato_results} for the binary decision task and the probabilistic decision task.

Corroborating with the authors' conclusion, by putting both tasks on the same payoff scale, we find that \textbf{people are better at the probability task than the decision task.}
First, we observe that the behavioral decision-makers doing the probability task can achieve higher performance than those doing the binary decision task overall. 
For example, \textbf{\textcolor{behavioral}{the behavioral performance}} for the probability task is much higher than \textbf{\textcolor{behavioral}{the behavioral performance}} for the binary decision task (Figure~\ref{fig:fogliato_results}).
Second, \textbf{\textcolor{behavioral}{the behavioral performance}} is higher than \textbf{\textcolor{baselinehuman}{the performance of the human only baseline}} in the probabilistic task while they
perform similarly in the decision task, as shown in Figure~\ref{fig:fogliato_results}B.
These results corroborate the conclusion by Fogliato et al. \cite{fogliato2021impact} that there is no determinstic decision rule that describes how the participants' probability estimates map to their binary decisions.

We also find that \textbf{the rational baselines and the rational benchmarks differ for each task between the anchoring and the no anchoring conditions, suggesting a need to reconsider Fogliato et al. \cite{fogliato2021impact}'s conclusion about the similarity between anchoring and no anchoring}. 
As shown in Figure~\ref{fig:fogliato_results}A, \textbf{\textcolor{baseline}{the rational baseline}} in the anchoring condition is slightly higher than in the non-achoring condition.
This implies just comparing the absolute performance of the behavioral decision-makers can mislead. 
Despite the behavioral performance being similar across the conditions in terms of absolute values, the behavioral decision-makers have better relative performance in the non-anchoring condition than the anchoring condition when compared to the rational baseline and the rational benchmark.

Similarly, contradicting the authors' conclusion, we find that \textbf{the behavioral decision-makers' reliance is closer to the appropriate reliance under the non-anchoring condition than the anchoring condition in both tasks.} 
As shown in Figure~\ref{fig:fogliato_results}C, the reliance loss ($\frac{\rdecision - \rmis}{\rdecision - \rprior}$) is lower for the no anchoring condition, while the discrimination loss ($\frac{\rmis - \baction}{\rdecision - \rprior}$) is slightly higher.
This suggests that letting the behavioral decision-makers make a decision by themselves first (a.k.a., the non-anchoring effects) can improve their reliance, but not necessarily help them distinguish between the signals where the AI recommendation is expected to outperform the human recommendation and the signals where the human recommendation is expected to outperform the AI recommendation.

\section{Discussion}

We contribute a formal definition of reliance and corresponding framework for interpreting losses in behavioral decision-making performance within the baseline and benchmark for complementary performance.
The first source of loss concerns the difference in the rate at which the behavioral decision-maker relies on the AI relative to the appropriate level of reliance defined by the decision problem, calculated in payoff space.
The second source of loss concerns the difference in score between a behavioral decision-maker and the best score a rational decision-maker who relies on the AI at the same rate as the behavioral decision-maker but who perfectly perceives the posterior probabilities could achieve.
By contributing clear comparison points in the form of performance \textit{benchmarks} to the design and interpretation of studies of human reliance on AI, our work enables researchers to identify the upper-bound of complementary performance and how far the human-AI team is from this optimal attainable performance.context

Closest to the motivation of our work, Fok and Weld \cite{fok2023search} motivate the need for a notion of ``strategy-graded reliance,'' where appropriate reliance is determined from the relative expected performance of the human and the AI, over ``outcome-graded reliance'' based on the human's acceptance of AI advice conditioned on its post-hoc correctness.
Several other empirical works propose studying reliance using conditional probability (e.g.,~\cite{schemmer2023appropriate,yang2020visual,wang2021explanations, schoeffer2023interdependence}) to separate cases where the human recommendation is better than the AI recommendation from cases where the AI recommendation is better than the human recommendation.
Information display in human-AI collaboration also has extended to using information of human accuracy before receiving AI~\cite{wang2022will, ma2023should}, suggesting studies about complementarity to consider correctness likelihood of both human and AI.
We unambigously define strategy-guided reliance and show how to calculate optimal reliance and disentangle sources of behavioral loss. 

Our framework enables evaluating reliance in payoff space, in contrast to prior work which has evaluated reliance in action space only~\cite{bansal2021does,schemmer2023appropriate,yang2020visual}.
Studying reliance only in the action space still neglects sensitivity in the payoff, such as the magnitude of improvement that the human recommendation provides over the AI recommendation or vice versa.
Defining a measurement of reliance in payoff space also enables the calculation of a benchmark to compare with, which we show in our demonstrations to be highly valuable for learning from a reliance evaluation.

Decoupling sources of behavioral loss in human AI-advised decisions is important for designing and interpreting AI-advised decision-making experiments, which helps to build better understanding and test hypothesis about the source of behavioral loss.
In recent years, numerous papers \cite{bansal2021does, yin2019understanding, feng2019can, fogliato2021impact, buccinca2021trust, chen2022use, hase2020evaluating, chen2023understanding, wang2021explanations, chouldechova2017fair, dressel2018accuracy, bussone2015role, yu2016trust, ashoori2019ai, hoff2015trust, horne2019rating, zhang2020effect,jiang2018trust, liu2021understanding, buccinca2020proxy, bansal2021does} have employed user studies to investigate how various factors
contribute to enhancing the complementary performance of human-AI teams.
Without a well-grounded notion of reliance, such studies have limited ability to draw conclusions from a decision-making task on how good the reliance is and whether action should be taken to improve it.
For example, in our demonstration of \citet{bansal2021does}, we find that the reliance level differing from optimal is not the main source of behavioral loss. This intepretation would suggest follow-up actions like calibrating human's trust on the AI in general (e.g., by making sure they have internalized information about its accuracy), but this may not adequately address challenges they face in discriminating which signals warrant accepting the AI's prediction. 
We also admit that while distinguishing reliance from discrimination loss in human-AI team performance may be useful to drive further improvements when there is a large discrepancy between these, in practice actions taken to improve one form of loss will likely affect the other.

Importantly, our framework hypothesizes two distinct roles in the decision-making process to separate human recommendations without AI assistance from the the process by which the human makes the final decision with access to human recommendations and AI recommendations.
This setup allows researchers to better interpret experiments and design the decision process they study; however, the generalizability of our framework to alternative study set-ups still holds. 
Our framework can be applied to situations where the human is both making a recommendation and making the final decision, i.e., where the human recommender and decision-maker are the same person. 
However, without constraints, they might ignore the AI and just submit the human recommendation or anchor on the AI without thinking to make the decision by themselves.
Both of these two cases cause inaccurate measurement of reliance, since AI recommendations and human recommendations are not consulted in human's decision rule.
Efforts should be made to align with the assumptions of our framework to facilitate the interpretation of experimental results.
\subsection{Limitations}

We formalize the AI-advised decision-making problem into a binary choice of whether to adopt a human recommendation or an AI recommendation.
However, this may not be suitable for every real world case. For example, when the recommendation space is continuous (e.g., regression), the human decision-maker is likely to make a decision that is different from the human recommendation or the AI recommendation.
Future work could extend our definition to continuous recommendation spaces.


We only identify two losses affecting human decision-makers, though more fine-grained losses may exist in AI-advised decision-making and be worth analyzing.
For example, discrimination loss can be caused by two possible reasons: misidentifying the probability that the AI is correct or misidentifying the probability that the human is correct. Improving the former implies better conveying the AI's accuracy, while improving the latter implies giving information on the human's average performance on the task.
More fine-grained behavioral losses can increase learning from experimental results and imply more targeted improvement of designs.
Future work can seek to identify and separate such additional behavioral losses and explore possible design choices to address them.


\begin{acks}
We would like to thank the authors of Lai and Tan \cite{lai2019human}, Bansal et al. \cite{bansal2021does}, and Fogliato et al. \cite{fogliato2021impact}, who provided their data for demonstration in this paper.
\end{acks}

\bibliographystyle{ACM-Reference-Format}
\bibliography{ref}

\clearpage

\appendix

\section{The algorithms for calculations in the framework}
\label{app:algorithms}

This appendix includes all the algorithms in the form of pseudocode for all the calculations we introduce in Section~\ref{sec:apply}.

\begin{algorithm}[hbt!]
\caption{Rational baseline}\label{alg:baseline}
\KwIn{the experimental data $D$ with each row representing one experimental trial, 
and the scoring rule for the derived binary-adoption decision task $\proper$}
\KwOut{the rational baseline $\rprior$}
$payoff \gets 0$\;
\For{$action \gets\{0, 1\}$ ($action=0$ follow human, $1$ follow AI)}{
\For{$row \in D$}{
  $\binadoptionstate \gets \text{the state realized in } row$\;
  $payoff \gets payoff + \proper(action, \binadoptionstate)$\;
}
$payoff_{action} \gets payoff / \text{the number of rows in }D$\;
}
$\rprior = \max\{payoff_0, payoff_1\}$\;
\end{algorithm}

\begin{algorithm}[hbt!]
\caption{Calculating the empirical distribution}\label{alg:overfit}
\KwIn{the experimental data $D$ with each row representing one experimental trial, the space of derived binary-adoption states $\binadoptionstatespace$, and the space of signals $\binadoptionsignalspace$}
\KwOut{the empirical distribution $\Tilde{\joint}(\binadoptionstate, \binadoptionsignal)$}
$\Tilde{\joint}(\binadoptionstate, \binadoptionsignal) \gets 0 \,{\mathds{1}}_{|\binadoptionstatespace|} {\mathds{1}}_{|\binadoptionsignalspace|}^\top$\; \Comment{Initializing a matrix with all $0$.}
\For{$row_i \in D$}{
  $\binadoptionstate_i \gets \text{the state realized in }row_i$\;
  $\binadoptionsignal_i \gets \text{the signal realized in }row_i$\;
  $\Tilde{\joint}(\binadoptionstate_i, \binadoptionsignal_i) \gets \Tilde{\joint}(\binadoptionstate_i, \binadoptionsignal_i) + 1$\;
}
$\Tilde{\joint}(\binadoptionstate, \binadoptionsignal) \gets \Tilde{\joint}(\binadoptionstate, \binadoptionsignal) / |\Tilde{\joint}(\binadoptionstate, \binadoptionsignal)|$\; \Comment{Normalizing to get the joint distribution.}
\end{algorithm}

\begin{algorithm}[hbt!]
\caption{Discretizing signals using the cluster generated by K-means}\label{alg:kmeans}
\KwIn{the experimental data $D$ with each row representing one experimental trial, the total number of clusters $K$, the space of derived binary-adoption states $\binadoptionstatespace$, and the space of signals $\signalspace$}
\KwOut{the empirical distribution $\Tilde{\joint}(\binadoptionstate, \Tilde{\signal})$ on the optimally discretized space}
$\Tilde{\joint}(\binadoptionstate, \Tilde{\signal}) \gets 0 \,{\mathds{1}}_{|\binadoptionstatespace|} {\mathds{1}}_{K}^\top$\; \Comment{Initializing a matrix with all $0$.}
$\{\binadoptionsignal_i\} \gets \text{all signals realized in }D$\;
$kmeans \gets initialize\_kmeans(\{\binadoptionsignal_i\}, K)$\;\Comment{Training the K-means model.}
\For{$row_i \in D$}{
  $\binadoptionstate_i \gets \text{the state realized in }row_i$\;
  $\binadoptionsignal_i \gets \text{the signal realized in }row_i$\;
  $\Tilde{\signal}_i \gets kmeans(\binadoptionsignal_i)$\;
  $\Tilde{\joint}(\binadoptionstate_i, \Tilde{\signal}_i) \gets \Tilde{\joint}(\binadoptionstate_i, \Tilde{\signal}_i) + 1$\;
}
$\Tilde{\joint}(\binadoptionstate, \Tilde{\signal}_i) \gets \Tilde{\joint}(\binadoptionstate, \Tilde{\signal}_i) / |\Tilde{\joint}(\binadoptionstate, \Tilde{\signal}_i)|$\; \Comment{Normalizing to get the joint distribution.}
\end{algorithm}

\begin{algorithm}[hbt!]
\caption{Rational benchmark}\label{alg:benchmark}
\KwIn{the experimental data $D$ with each row representing one experimental trial, the joint distribution between states and signals $\joint(\binadoptionstate, \binadoptionsignal)$, and the scoring rule for the derived binary-adoption decision task $\proper$}
\KwOut{the rational benchmark $\rdecision$}
$payoff \gets 0$\;
\For{$row \in D$}{
  $\binadoptionsignal \gets \text{the signal realized in } row$\; 
  $\joint(\binadoptionstate|\binadoptionsignal) = \joint(\binadoptionstate, \binadoptionsignal) / \joint(\binadoptionsignal)$\;\Comment{the posterior distribution of the binary-adoption state}
  $action \gets \argmax_{\binadoptionaction \sim \{\text{human}, \text{AI}\}}{E_{\binadoptionstate \sim \joint(\binadoptionstate|\binadoptionsignal)}(\proper(\binadoptionaction, \theta))}$\;\Comment{the action made on the posterior distribution}
  $\binadoptionstate \gets \text{the state realized in } row$\;
  $payoff \gets payoff + \proper(action, \binadoptionstate)$\;
}
$\rdecision \gets payoff / \text{the number of row in }D$\;
\end{algorithm}

\begin{algorithm}[hbt!]
\caption{Behavioral performance}\label{alg:behavioral}
\KwIn{the experimental data $D$ with each row representing one experimental trial, the joint behavior $\joint(\signal, \payoffstate, \actb)$, and the scoring rule $\score$}
\KwOut{the behavioral performance $\baction$}
$payoff \gets 0$\;
\For{$row \in D$}{
  $\signal \gets \text{the signal realized in } row$\; 
  $\payoffstate \gets \text{the state realized in } row$\;
  $action \gets \text{action drawn from } \joint(\actb | \payoffstate, \signal) = \joint(\signal, \payoffstate, \actb) / \joint(\payoffstate, \signal)$\;
  $payoff \gets payoff + \score(action, \payoffstate))$\;
}
$\baction \gets payoff / \text{the number of row in }D$\;
\end{algorithm}

\begin{algorithm}[hbt!]
\caption{Mis-reliant rational benchmark}\label{alg:misreliant}
\KwIn{the experimental data $D$ with each row representing one experimental trial, the joint distribution between states and signals $\joint(\binadoptionstate, \binadoptionsignal)$, the scoring rule for the original decision task $\score$, and the scoring rule for the derived binary-adoption decision task $\proper$}
\KwOut{the mis-reliant rational benchmark $\rmis$}
$P \gets \{P_1, \ldots, P_M\}$\; \Comment{The sets of trials finished by each participant; $M$ participants in total.}
\For{$i \in \{1, \ldots, M\}$}{
    $P_i \gets filter(D, participant\_id == i)$\;
}
$payoff \gets 0$\;
\For{$P_i \in P$}{
  $\text{Sort }P_i\text{ in decreasing order of }\expect[\joint(\binadoptionstate | \binadoptionsignal)]{\score(\aipred, \pred)} - \expect[\joint(\binadoptionstate | \binadoptionsignal)]{\score(\hpred, \pred)}$\;
  $\{\binadoptionsignal_j\} \gets \{\text{the signal realized in }row_j\}_{row_j \in P_i}$\;
  $\{\payoffstate_j\} \gets \{\text{the state realized in }row_j\}_{row_j \in P_i}$\;
  $\{\actb_j\} \gets \{\text{action drawn from } \joint(\actb | \payoffstate_j, \binadoptionsignal_j)\}_{row_j \in P_i}$\;
  $\rl^b \gets \sum_{row_j \in P_i}{\mathds{1}[\actb_j = \aipred_j \& \aipred_j \neq \hpred_j]}$\;
  $N \gets \text{the number of rows in }P_i$\;
  $\{\actr_j\} \gets \{AI\}_{j \in \{1, \ldots, \rl^b\}} \cup \{human\}_{j \in \{\rl^b + 1, \ldots, N\}}$\;
  $\{\binadoptionstate_j\} \gets \{\text{the binary-adoption state realized in }row_j\}_{row_j \in P_i}$\;
  $payoff \gets payoff + \sum_{j \in [N]} \score(\actr_j, \binadoptionstate_j))$\;
}
$\rmis \gets payoff / \text{the number of rows in }D$\;
\end{algorithm}

\begin{algorithm}[hbt!]
\caption{Quantifying uncertainty}\label{alg:uncertainty}
\KwIn{the experimental data $D$ with each row representing one experimental trial, total number of iterations $T$, the sample size $k$, prior distribution of the binary-adoption state $\joint(\binadoptionstate)$,  the joint distribution between states and signals $\joint(\binadoptionstate, \binadoptionsignal)$, the joint behavior $\joint(\signal, \payoffstate, \actb)$, the scoring rule $\score$, and the scoring rule for derived binary-adoption decision task $\proper$}
\KwOut{the distribution of the rational baseline $\{\rprior{}_i\}_{i\in [T]}$, the rational benchmark $\{\rdecision_i\}_{i \in [T]}$, the behavioral performance $\{\baction_i\}_{i\in [T]}$, and the mis-reliant rational benchmark $\{\rmis_i\}_{i\in [T]}$}
\For{$i \in [T]$}{
  $\Tilde{D} \gets sample(D, k)$\;
  $\rprior{}_i \gets \text{Rational baseline}(\Tilde{D}, \joint(\binadoptionstate), \proper)$\;
  $\rdecision_i \gets \text{Rational benchmark}(\Tilde{D}, \joint(\binadoptionstate, \binadoptionsignal), \proper)$\;
  $\baction_i \gets \text{Behavioral performance}(\Tilde{D}, \joint(\signal, \payoffstate, \actb), \score)$\;
  $\rmis_i \gets \text{Mis-reliant rational baseline}(\Tilde{D}, \joint(\binadoptionstate, \binadoptionsignal), \score, \proper)$\;
}
\end{algorithm}

\clearpage

\newpage

\section{Formalized decision tasks}
\label{app:decision_task}

This appendix includes the formalized decision tasks for the experiments in the main text.

\begin{table*}[htbp]
\small
    \centering
    \begin{tabular}{c|l|l}
    \hline
    & The original decision task & The derived binary-adoption decision task \\
    \hline
     Payoff-related state   &$\payoffstate = \text{Ground truth } \pred \in \{0, 1\}$ &  $\binadoptionstate = (\pred, \hpred, \aipred)$ \\
     &\quad Be re-arrested or not&  Ground truth $\pred\in \{0, 1\}$ \\
     && Human recommendation $\hpred \in \{0, 1\}$ \\
     && AI recommendation $\aipred \in \{0, 1\}$\\
     \hline
     Data generating model    
  & \multicolumn{2}{|l}{A profile $\feature$ of a defendent who is randomly drawn from the defendent population}\\
 &\multicolumn{2}{|l}{Ground truth $\pred$ drawn from a distribution conditioned on $x$.}\\
 &\multicolumn{2}{|l}{The human recommendation $\hpred$ is produced by the decision rule of the human predictor,}\\
 &\multicolumn{2}{|l}{\ \ represented by the joint behavioral $\joint(\hpred, \feature, \pred)$}\\
 &\multicolumn{2}{|l}{AI recommendation $\aipred$ for the profile $\feature$}\\
 &\multicolumn{2}{|l}{The explanation $\explanation(\aipred)$}\\
         \hline
         Action (choice) & $\action \in \{0, 1\}$ Be re-arrested or not & $\binadoptionaction\in \{0 = \text{human}, 1 = \text{AI}\}$\\
     \hline 
     Signal  & $\signal = \{\feature, \aipred, \explanation(\aipred)\}$ & $\binadoptionsignal = \{\feature, \hpred, \aipred\}$\\
     \hline
     Scoring rule (payoff) & $\score(\action, \payoffstate) = 0.5 \times \mathds{1}[\action = \payoffstate]$  & $\proper(\binadoptionaction, \binadoptionstate)=\score(\hpred, \pred)$ if $\binadoptionaction=\text{human}$\\
& &$\proper(\binadoptionaction, \binadoptionstate)=\score(\aipred, \pred)$ if $\binadoptionaction=\text{AI}$
     \\
     \hline
    \end{tabular}
    \caption{Example of original and derived binary-adoption decision task in hypothetical recidivism experiment}
    \normalsize
    \label{tab:applying_example}
\end{table*}

\begin{table*}[htbp]
\small
    \centering
    \begin{tabular}{c|l|l}
    \hline
     Payoff-related state    &  $\payoffstate = \text{Correct answer } \pred \in \{A, B, C, D\} $  & $\binadoptionstate = (\pred, \hpred, \aipred)$ \\
 & & Ground truth $\pred\in \{A, B, C, D\}$ \\
 & &Human recommendation $\hpred \in \{A, B, C, D\}$ \\
 & &AI recommendation $\aipred \in  \{A, B, C, D\}$\\
     \hline
     Data generating model 
  & \multicolumn{2}{l}{Question $\feature$ drawn from the scope of LSAT questions}\\
 &\multicolumn{2}{l}{Correct answer $\pred$ for $\feature$}\\
 &\multicolumn{2}{l}{AI recommendation $\aipred$ for $\feature$}\\
 &\multicolumn{2}{l}{Human recommendation $\hpred$:}\\
 &\multicolumn{2}{l}{\quad $\hpred \sim \joint(\feature, \hpred)/\joint(\feature)$}\\
 &\multicolumn{2}{l}{Explanation $\explanation(\aipred)$}\\
         \hline
         Action (choice) & $\action\in  \{A, B, C, D\}$ &$\action\in  \{0 = \text{human}, 1 = \text{AI}\}$\\
     \hline 
     Signal  & $\signal = \{\feature, \aipred, \explanation(\aipred)\}$ & $\binadoptionsignal = \{\feature, \hpred, \aipred\}$\\
     \hline
     Scoring rule (payoff) & $\score(\action, \payoffstate) = \mathds{1}[\action = \payoffstate] $
      &$\proper(\binadoptionaction, \binadoptionstate)=\score(\hpred, \pred)$ if $\binadoptionaction=\text{human}$\\
      &&$\proper(\binadoptionaction, \binadoptionstate)=\score(\aipred, \pred)$ if $\binadoptionaction=\text{AI}$\\
     \hline
    \end{tabular}
    \caption{\citet{bansal2021does} decicion task under our framework.}
    \normalsize
    \label{tab:notation_bansal}
\end{table*}

\begin{table*}[!htbp]
\small
    \centering
    \begin{tabular}{c|l|l}
    \hline
    & The original decision task & The derived binary-adoption decision task \\
    \hline
     Payoff-related state   &$\payoffstate = \text{Ground truth } \pred \in \{0,1\}$&  $\binadoptionstate = (\pred, \hpred, \aipred)$ \\
     &\quad Deceptive or genuine&  Ground truth $\pred\in \{0,1\}$\\
     && Human recommendation $\hpred \in \{0,1\}$\\
     && AI recommendation $\aipred \in \{0,1\}$\\
     \hline
     Data generating model    
  & \multicolumn{2}{|l}{Ground truth $\pred \sim \text{Bernoulli}(0.5)$, indicating whether the review}\\
  & \multicolumn{2}{|l}{\ \ is written by a person who has been going to the hotel or not.}\\
  & \multicolumn{2}{|l}{Review text $\feature$ generated by the person}\\
  & \multicolumn{2}{|l}{\quad $\feature \sim \joint(\feature, \pred)/\joint(\pred)$}\\
 &\multicolumn{2}{|l}{Human recommendation $\hpred$}\\
 & \multicolumn{2}{l}{\quad $\hpred \sim \joint(\feature, \hpred) / \joint(\feature)$}\\
 &\multicolumn{2}{|l}{AI recommendation $\aipred$ for $\feature$}\\
 & \multicolumn{2}{|l}{Explanation $\explanation(\aipred)$}\\
         \hline
         Action (choice) & $\action \in \{0,1\}$ Deceptive or genuine& $\binadoptionaction\in \{0 = \text{human}, 1 = \text{AI}\}$\\
     \hline 

     Signal  & $\signal = \{\feature, \aipred, \explanation(\aipred)\}$ & $\binadoptionsignal = \{\feature, \hpred, \aipred\}$\\
     \hline
     Scoring rule (payoff) & $\score(\action, \payoffstate) = \mathds{1}[\action = \payoffstate]$  & $\proper(\binadoptionaction, \binadoptionstate)=\score(\hpred, \pred)$ if $\binadoptionaction=\text{human}$\\
& &$\proper(\binadoptionaction, \binadoptionstate)=\score(\aipred, \pred)$ if $\binadoptionaction=\text{AI}$\\
     
     \hline\end{tabular}
    \caption{\citet{lai2019human} decision task under our framework.}
    \normalsize
    \label{tab:lai_decision}
\end{table*}

Note that the feature $\feature$ is included in the binary-adoption signals in \Cref{tab:lai_decision} because $\hpred$ and $\aipred$ in \citet{lai2019human} are both binary-valued, which makes the signal space too small to generate a meaningful benchmark.

\begin{table*}[htbp]
\small
    \centering
    \begin{tabular}{c|l|l|l}
    \hline
    & The binary decision task &The probabilistic decision task& The binary-adoption decision task \\
    \hline
     Payoff-related state   &\multicolumn{2}{l|}{ $\payoffstate = \text{Ground truth }\pred \in \{0, 1\}$ Re-arrest or not}&  $\binadoptionstate = (\pred, \hpred, \aipred)$ \\
     &\multicolumn{2}{l|}{}&  Ground truth $\pred\in \{0,1\}$\\
     & \multicolumn{2}{l|}{}& Human recommendation $\hpred \in \{0,1\}$\\
     & \multicolumn{2}{l|}{}& AI recommendation $\aipred \in \{0,1\}$\\
     \hline
     Data generating model    
  & \multicolumn{3}{|l}{A defendent $p$ is randomly drawn from the defendent population.} \\
  & \multicolumn{3}{|l}{The profile $\feature$ for $p$} \\
 &\multicolumn{3}{|l}{Ground truth $\pred$ for $p$} \\
 &\multicolumn{3}{|l}{Human recommendation $\hpred$} \\
 & \multicolumn{3}{l}{\quad $\hpred \sim \joint(\feature, \hpred)/\joint(\feature)$} \\
 &\multicolumn{3}{|l}{AI recommendation $\aipred$ for $\feature$} \\
 & \multicolumn{3}{|l}{AI's confidence score $\explanation(\aipred)$} \\
         \hline
         Action (choice) & $\action \in \{0,1\}$ Re-arrest or not&$\action \in [0, 1]$ Probability of re-arrest& $\binadoptionaction\in \{0 = \text{human}, 1 = \text{AI}\}$\\
     \hline 

     Signal  & \multicolumn{2}{l|}{$\signal = \{\feature, \aipred, \explanation(\aipred)\}$} & $\binadoptionsignal = \{\feature, \hpred, \aipred\}$\\
     \hline
     Scoring rule (payoff) & \multicolumn{2}{l|}{$\score(\action, \payoffstate) = 1 - (\action - \payoffstate)^2$}& $\proper(\binadoptionaction, \binadoptionstate)=\score(\hpred, \pred)$ if $\binadoptionaction=\text{human}$\\
&  \multicolumn{2}{l|}{}&$\proper(\binadoptionaction, \binadoptionstate)=\score(\aipred, \pred)$ if $\binadoptionaction=\text{AI}$\\
     
     \hline\end{tabular}
    \caption{\citet{fogliato2021impact} decision task under our framework.}
    \normalsize
    \label{tab:notation_fogliato}
\end{table*}

\clearpage

\newpage

\section{The results of demonstrations using discretized signal approximation}
\label{app:demo_test_performance}

This appendix includes our additional results for demonstrations in Section~\ref{sec:demonstration}, where we use the discretized signals to approximate the rational benchmark and the mis-reliant rational benchmark.
We subsequently re-check the conclusions we get in Section~\ref{sec:demonstration} with the results shown in this appendix.
All the conclusions analyzed under the results of approximation using discretized signals corroborate with the conclusions we get in Section~\ref{sec:demonstration}.

\subsection{Does the Whole Exceed its Parts?~\cite{bansal2021does}}

\begin{figure}[!h]
    \centering
    \includegraphics[width=\linewidth]{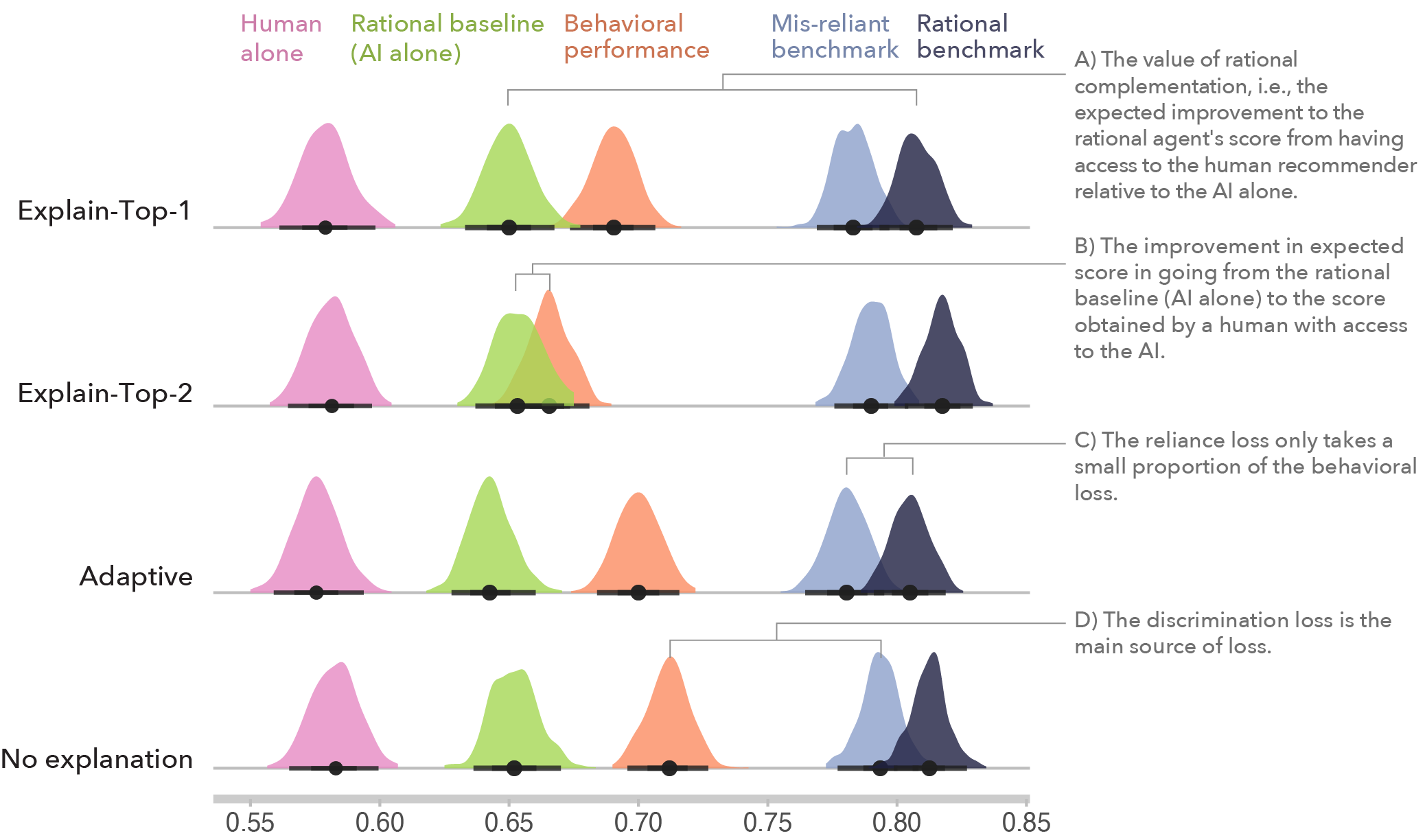}
    \caption{Estimated payoffs of the experiment data in \citet{bansal2021does}.}
    \label{fig:bansal_results_test_performance}
\end{figure}

First, the results also show considerable room for improvement to achieve to \textbf{\textcolor{benchmark}{the rational benchmark}}, as shown in Figure~\ref{fig:bansal_results_test_performance}A and B.
Second, no significant improvement by displaying explanations is evidenced in the results.
As shown by Figure~\ref{fig:bansal_results_test_performance}, the \textbf{\textcolor{behavioral}{behavioral performance}} and the\textbf{\textcolor{misreliant}{mis-reliant rational benchmark}} perform similarly across the explanation conditions and the no explantion condition.
Third, the reliance loss is modest to the behavioral loss, while the discrimination loss is the main source of loss, as shown in Figure~\ref{fig:bansal_results_test_performance}C and D.

\subsection{On Human Predictions with Explanations and Predictions of Machine Learning Models~\cite{lai2019human}}

\begin{figure}[!h]
    \centering
    \includegraphics[width=\linewidth]{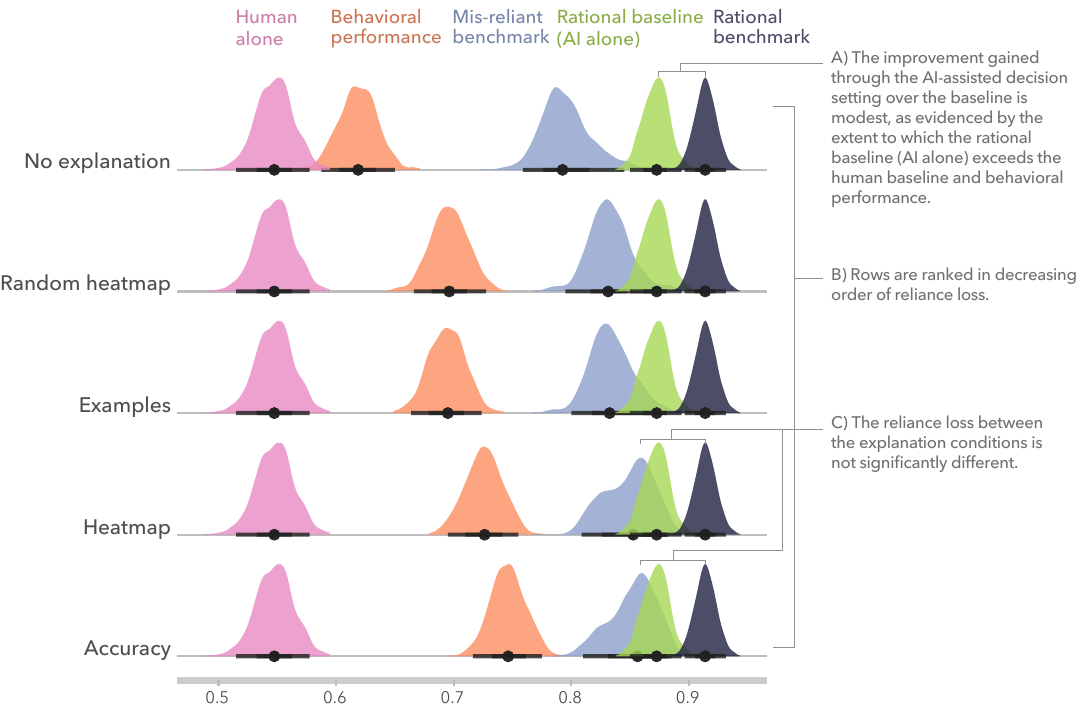}
    \caption{Estimated payoffs of the experiment data in \citet{lai2019human}.}
    \label{fig:lai_results_test_performance}
\end{figure}

First, similarly to what we get in Section~\ref{sec:demonstration}, \textbf{\textcolor{baseline}{the rational baseline}} dominates all other quantities defined by our framework except \textbf{\textcolor{benchmark}{the rational benchmark}}, leading to the conclusion about the failure of complementary performance in the decision task.
Second, \textbf{\textcolor{benchmark}{the rational benchmark}} only shows marginal improvement over \textbf{\textcolor{baseline}{the rational baseline}}, as shown in Figure~\ref{fig:bansal_results_test_performance}A.
Third, the explanations can improve the behavioral performance and the reliance, as shown in Figure~\ref{fig:bansal_results_test_performance}C.
Finally, we observed the same pattern of reliance loss and discrimination loss in the results, e.g., Figure~\ref{fig:bansal_results_test_performance}D.

\subsection{The Impact of Algorithmic Risk Assessments on Human Predictions and its Analysis via Crowdsourcing Studies~\cite{fogliato2021impact}}

\begin{figure}[!h]
    \centering
    \includegraphics[width=\linewidth]{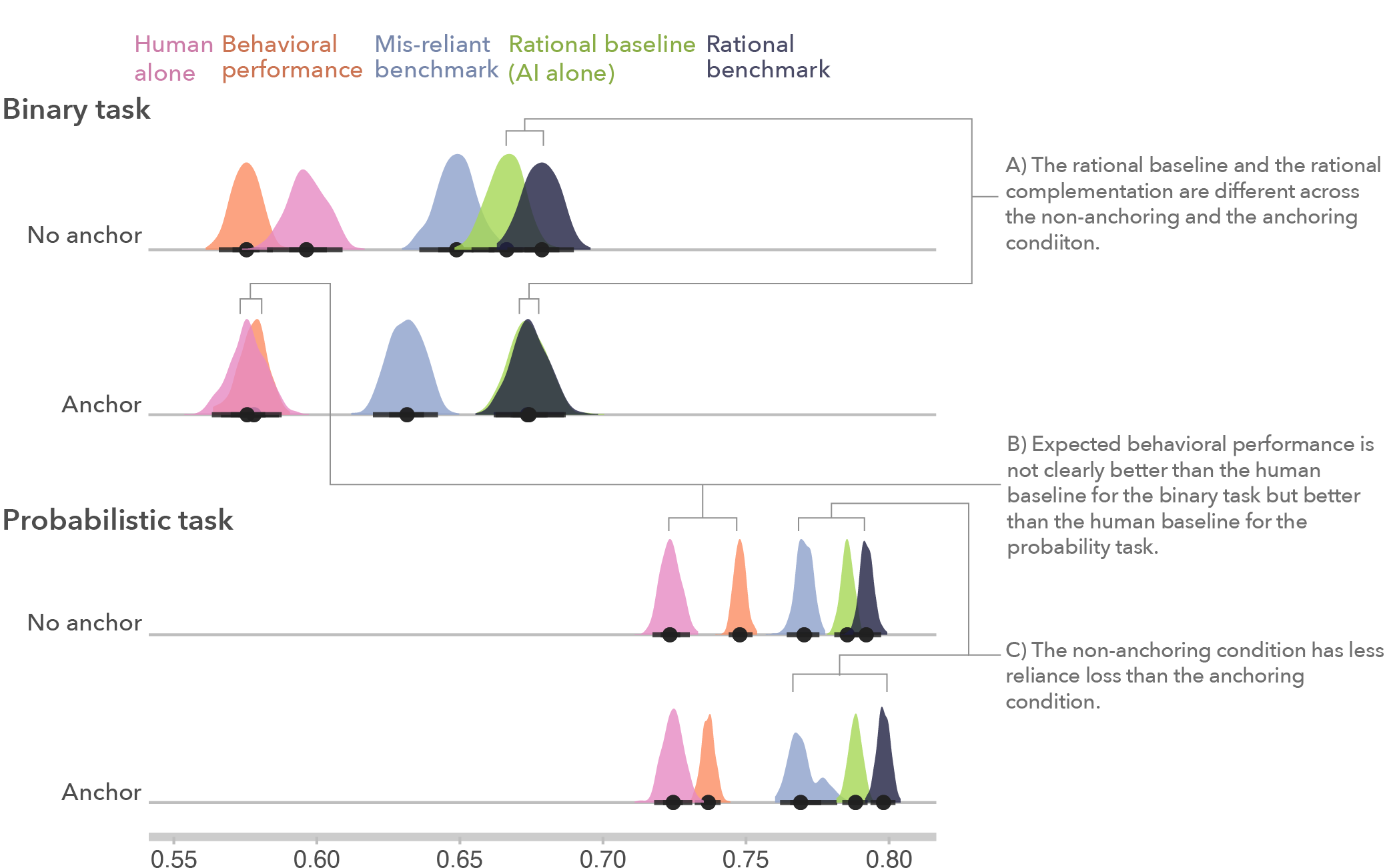}
    \caption{Estimated payoffs of the experiment data in \citet{fogliato2021impact}.}
    \label{fig:fogliato_results_test_performance}
\end{figure}

First, we also find the quantities under our framework act differently between the probabilistic decision task the the binary decision task.
For example, \textbf{\textcolor{behavioral}{the behavioral performance}} exceeds \textbf{\textcolor{baselinehuman}{the performance of human predictions}} in the probabilistic decision task while acts the same in the binary decision task (Figure~\ref{fig:bansal_results_test_performance}B).
Second, \textbf{\textcolor{baseline}{the rational baseline}} and \textbf{\textcolor{benchmark}{the rational benchmark}} have different values on the anchoring effect condition and the non-anchoring effect condition, as shown in Figure~\ref{fig:bansal_results_test_performance}A.
Finally, the anchoring effect condition can improve the reliance loss over the non-anchoring effect condition, as shown in Figure~\ref{fig:bansal_results_test_performance}C.

\end{document}